\documentclass[10pt,twocolumn,letterpaper]{article}

\usepackage{cvpr}
\makeatletter
\@namedef{ver@everyshi.sty}{}
\makeatother

\usepackage{times}
\usepackage{epsfig}
\usepackage{graphicx}
\usepackage{amsmath}
\usepackage{amssymb}

\usepackage{amsmath,amsfonts,bm,amssymb}

\def\eqref#1{equation~\ref{#1}}

\def\1{\bm{1}}

\DeclareMathAlphabet{\mathsfit}{\encodingdefault}{\sfdefault}{m}{sl}
\SetMathAlphabet{\mathsfit}{bold}{\encodingdefault}{\sfdefault}{bx}{n}

\newcommand{\red}[1]{\textcolor{red}{#1}}

\newcommand{\smallpar}[1]{\smallskip\noindent {\bf{#1}}}

\newcommand{\DA}{$\downarrow$\@}

\newcommand{\tih}{\textit{tieredImageNet-H}}
\newcommand{\inh}{\textit{iNaturalist-19}}

\DeclareRobustCommand{\hlyellow}[1]{{\sethlcolor{yellow}\hl{#1}}}
\DeclareRobustCommand{\hlcyan}[1]{{\sethlcolor{lightgray}\hl{#1}}}

\newcommand{\devise}{DeViSE{}}
\newcommand{\BD}{Barz\&Denzler{}}
\newcommand{\yolo}{YOLO-v2{}}

\newenvironment{tight_it}{
\begin{itemize}
    \setlength{\itemsep}{0pt}
    \setlength{\parskip}{0pt}
    \setlength{\parsep}{0pt}
}{\end{itemize}}

\newenvironment{tight_enum}{
\begin{enumerate}
		\setlength{\itemsep}{0pt}
		\setlength{\parskip}{0pt}
		\setlength{\parsep}{0pt}
}{\end{enumerate}}

\usepackage{url}
\usepackage{graphicx}
\usepackage{float}
\usepackage{subfig}
\usepackage{forest}
\usepackage{booktabs}
\usepackage{soul}
\usepackage{siunitx}
\usepackage{colortbl}
\usepackage{listings}
\usepackage[pagebackref=true,breaklinks=true,letterpaper=true,colorlinks,bookmarks=false]{hyperref}

\cvprfinalcopy 

\def\cvprPaperID{} 

\ifcvprfinal\fi
\begin{document}

\title{Making Better Mistakes: Leveraging Class Hierarchies with Deep Networks}
\thispagestyle{plain}
\pagestyle{plain}

\author{
Luca Bertinetto\thanks{Equal contribution.}
\quad Romain Mueller$^*$
\quad Konstantinos Tertikas
\quad Sina Samangooei
\quad Nicholas A. Lord$^*$
\\
{\tt\small \{luca.bertinetto, romain.mueller, konstantinos.tertikas, sina, nick.lord\}@five.ai}\\
\url{www.five.ai}\\
}

\maketitle

\begin{abstract}
Deep neural networks have improved image classification dramatically over the past decade, but have done so by focusing on performance measures that treat all classes other than the ground truth as equally wrong.
This has led to a situation in which mistakes are less likely to be made than before, but are equally likely to be absurd or catastrophic when they do occur.
Past works have recognised and tried to address this issue of mistake severity, often by using graph distances in class hierarchies, but this has largely been neglected since the advent of the current deep learning era in computer vision. 
In this paper, we aim to renew interest in this problem by reviewing past approaches and proposing two simple modifications of the cross-entropy loss which outperform the prior art under several metrics on two large datasets with complex class hierarchies: tieredImageNet and iNaturalist'19.
\end{abstract}

\vspace{-0.4cm}
\section{Introduction}
\label{sec:introduction}
Image classification networks have improved greatly over recent years, but generalisation remains imperfect, and test-time errors do of course occur.
Conventionally, such errors are defined with respect to a single ground-truth class and reported using one or more top-$k$ measures ($k$ typically set to $1$ or $5$).
However, this practice imposes certain notions of what it means to make a mistake, including treating all classes other than the ``true" label as equally wrong.
This may not actually correspond to our intuitions about desired classifier behaviour, and for some applications this point may prove crucial.
Take the example of an autonomous vehicle observing an object on the side of the road: whatever measure of classifier performance we use, we can certainly agree that mistaking a lamppost for a tree is less of a problem than mistaking a person for a tree, as such a mistake would have crucial implications in terms both of prediction and planning. 
If we want to take such considerations into account, we must incorporate a nontrivial model of the relationships between classes, and accordingly rethink more broadly what it means for a network to ``make a mistake".
One natural and convenient way of representing these class relationships is through a taxonomic hierarchy tree.

\begin{figure}[t]
    \centering
    \includegraphics[width=0.5\textwidth]{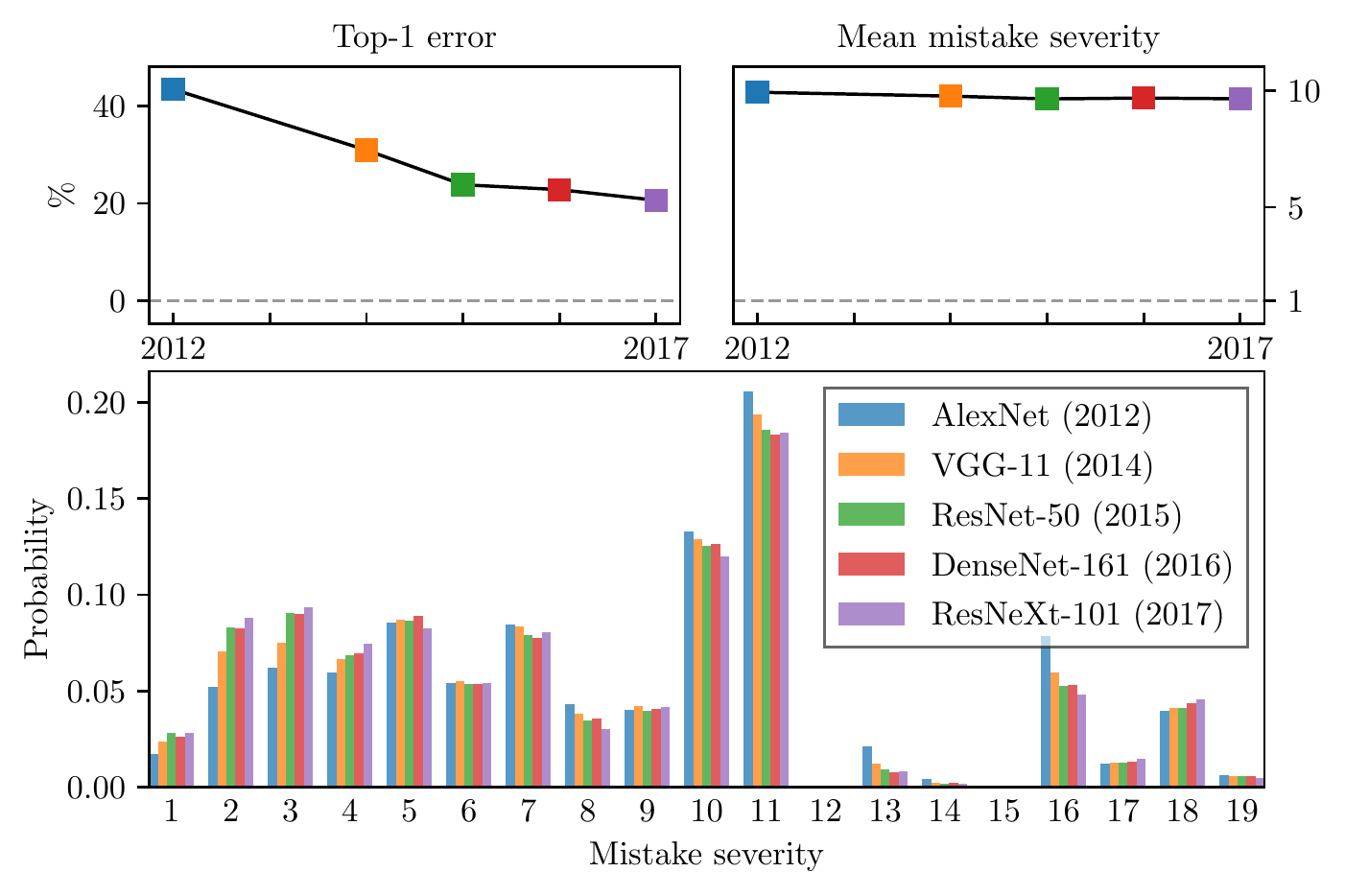}
    \vspace{-0.6cm}
    \caption{\label{fig:history}
		Top-1 error and distribution of mistakes w.r.t. the WordNet hierarchy for well-known deep neural network architectures on ImageNet/ILSVRC-12: see text for definition of mistake severity.
		The top-1 error has seen a spectacular improvement in the last few years, but though the \emph{number} of mistakes has decreased in absolute terms, the \emph{severity} of the mistakes made has remained fairly unchanged.
Dashed lines denote the best achievable value of each metric.
}
\end{figure}

This idea is not new.
In fact, it was once fairly common across various machine learning application domains to consider class hierarchy when designing classifiers, as surveyed in Silla \& Freitas~\cite{silla2011survey}.
That work assembled and categorised a large collection of hierarchical classification problems and algorithms, and suggested widely applicable measures for quantifying classifier performance in the context of a given class hierarchy.
The authors noted that the hierarchy-informed classifiers of the era typically empirically outperformed ``flat" (\ie~hierarchy-agnostic) classifiers even under standard metrics, with the performance gap increasing further under the suggested hierarchical metrics.
Furthermore, class hierarchy is at the core of the ImageNet dataset: as detailed in Deng~\etal~\cite{deng2009imagenet}, it was constructed directly from WordNet~\cite{miller1998wordnet}, itself a hierarchy originally designed solely to represent semantic relationships between words.
Shortly after ImageNet's introduction, works such as Deng~\etal~\cite{deng2010does}, Zhao~\etal~\cite{zhao2011large}, and Verma~\etal~\cite{verma2012learning} explicitly noted that the underpinning WordNet hierarchy suggested a way of quantifying the severity of mistakes, and experimented with hierarchical cost minimisation.
Likewise, Deng~\etal~\cite{deng2011hierarchical} presented a straightforward method for using a hierarchy-derived similarity matrix to define a more semantically meaningful compatibility function for image retrieval.
Despite this initial surge of interest and the promising results accompanying it, the community effectively discarded hierarchical measures after deciding that they were not communicating substantially different information about classifier performance than top-1 and top-5 accuracies\footnote{From Russakovsky~\etal~\cite{russakovsky2015imagenet}: \textit{``[..] we found that all three measures of error (top-5, top-1, and hierarchical) produced the same ordering of results.
Thus, since ILSVRC2012 we have been exclusively using the top-5 metric which is the simplest and most suitable to the dataset.''}}.
When the celebrated results in Krizhevsky~\etal~\cite{krizhevsky2012imagenet} were reported in flat top-$k$ terms only, the precedent was firmly set for the work which followed in the deep learning era of image classification.
Interest in optimising hierarchical performance measures waned accordingly.

We argue here that this problem is ripe for revisitation, and we begin by pointing to Fig.~\ref{fig:history}.
Here, a \emph{mistake} is defined as a top-1 prediction which differs from the ground-truth class, and the \emph{severity} of such a mistake is the height of the lowest common ancestor of the predicted and ground-truth classes in the hierarchy.
We see that while the flat top-1 accuracies of state-of-the-art classifiers have improved to impressive levels over the years, the distributions of the severities of the errors that \emph{are} made have changed very \emph{little} over this time.
We hypothesise that this is due, at least in part, to the scarcity of modern learning methods which attempt to exploit prior information and preferences about class relationships in the interest of ``making better mistakes", whether this information is sourced from an offline taxonomy or otherwise.
The few exceptions of which we are aware include Frome~\etal~\cite{frome2013devise}, Wu~\etal\cite{wu2016learning},
Barz \& Denzler~\cite{barz2019hierarchy}, and a passing mention in Redmon \& Farhadi~\cite{redmon2017yolo9000}.
In Sec.~\ref{sec:related}, we suggest a framework for thinking about these pieces of work, their predecessors, and some of their conceptual relatives.

The contributions of this work are as follows:
\begin{tight_enum}
	\item We review relevant literature within an explanatory framework which unifies a fairly disjoint prior art.
	\item Building on the perspective gained from the preceding, we propose two methods that are both simple and effective at leveraging class hierarchies.
Each uses a one-parameter drop-in generalisation of the standard cross-entropy loss.
These loss variants can be tuned to produce different empirical tradeoffs between top-$k$ and hierarchical performance measures, and reduce to the standard setup in their respective limits.
	\item We perform an extensive experimental evaluation to both demonstrate the effectiveness of the said methods compared to prior art and to encourage future work.
\end{tight_enum}
The PyTorch~\cite{paszke2019pytorch} code for all experiments is available at \href{https://github.com/fiveai/making-better-mistakes}{github.com/fiveai/making-better-mistakes}.

\section{Framework and related work}
\label{sec:related}

We first suggest a simple framework for thinking about methods relevant to the problem of making better mistakes on image classification, beginning with the standard supervised setup.
Consider a training set $\mathcal S = \{(x_i, C_i)\}_{i=1, \ldots, N}$ which pairs $N$ images $x_i \in \mathcal I$ with class labels $C_i \in \mathcal C$.
A network architecture implements the predictor function $\phi(x; \theta)$, whose parameters $\theta$ are learned by minimising
\begin{equation} \label{eq:risk}
\frac 1 N \sum_{i=1}^N \mathcal{L}(\phi(x_i; \theta), y(C_i)) + \mathcal{R}(\theta),
\end{equation}
where the loss function $\mathcal L$ compares the predictor's output $\phi(x_i; \theta)$ to an embedded representation $y(C_i)$ of each example's class, and $\mathcal R$ is a regulariser.

Under common choices such as cross-entropy for $\mathcal L$ and one-hot embedding for $y$, it is 
easy to see that the framework is agnostic of relationships between classes.
The question is how such class relationships $\mathcal H$
can be incorporated into the loss in Eqn.~\ref{eq:risk}.
We identify the following three approaches: 
\begin{tight_enum}
	\item Replacing class representation $y(C)$ with an alternate embedding $y^{\mathcal H}(C)$.
	Such ``label-embedding" methods, discussed in Sec.~\ref{subsec:soft_targets_embeddings}, can draw their embedding both from taxonomic hierarchies and alternative sources.
	
	\item Altering the loss function $\mathcal L$ in terms of its arguments to produce $\mathcal{L}^{\mathcal H}(\phi(x; \theta), y(C))$,~\ie~making the penalty assigned to a given output distribution and embedded label dependent on $\mathcal H$.
	Methods using these
	``hierarchical losses" 
	are covered in Sec.~\ref{subsec:explicit_hierarchical_output_penalty}.
	
	\item Altering the function $\phi(x; \theta)$ to $\phi^{\mathcal H}(x; \theta)$,
	\ie~making hierarchically-informed architectural changes to the network, generally with the hope of introducing a favourable inductive bias.
	We cover these ``hierarchical architectures" in Sec.~\ref{subsec:divide_and_conquer} .
\end{tight_enum}
While a regulariser $\mathcal{R}^\mathcal{H}$ is certainly feasible, it is curiously rare in practice:~\cite{zhao2011large} is the only example we know of.

\subsection{Label-embedding methods}
\label{subsec:soft_targets_embeddings}

These methods map class labels to vectors whose relative locations represent semantic relationships, and optimise a loss on these embedded vectors.
The DeViSE method of Frome~\etal~\cite{frome2013devise} maps target classes onto a unit hypersphere,
assigning terms with similar contexts to similar representations through analysis of unannotated Wikipedia text~\cite{mikolov2013efficient}.
The loss function is a ranking loss which penalises the extent to which the output is more cosine-similar to false label embeddings than to the correct one.
They learn a linear mapping from a pre-trained visual feature pipeline
to the embedded labels, then fine-tune the visual pipeline.

Romera-Paredes \& Torr~\cite{romera2015embarrassingly} note that their solution for learning an analogous linear mapping for zero-shot classification should easily extend to accommodating these sorts of embeddings.
In Hinton~\etal~\cite[Sec.~2]{hinton2015distilling}, the role of the label embedding function is played by a temperature-scaled pre-existing classifier ensemble.
This ensemble is  ``distilled" into a smaller DNN through cross-entropy minimisation against the ensemble's output.
For zero-shot classification, Xian~\etal~\cite{xian2016latent} experiment with various independent embedding methods, as is also done in Akata~\etal~\cite{akata2015evaluation}: annotated attributes, word2vec~\cite{mikolov2013distributed}, glove~\cite{pennington2014glove}, and the WordNet hierarchy.
Their ranking loss function is functionally equivalent to that in Frome~\etal~\cite{frome2013devise}, and they learn a choice of linear mappings to these representations from the output of a fixed CNN.
Barz \& Denzler~\cite{barz2019hierarchy} present an embedding algorithm which maps examples onto a hypersphere such that all
distances represent similarities derived from lowest common ancestor (LCA) height in a given hierarchy tree.
They proceed by minimising the sum of two rather different losses:
(1) a linear loss based on cosine distance to the embedded class vectors
and (2) the standard cross-entropy loss on the output of a fully-connected layer added after the embedding layer.

\subsection{Hierarchical losses}
\label{subsec:explicit_hierarchical_output_penalty}

In these methods, the loss function itself is parametrised by the class hierarchy such that a higher penalty is assigned to the prediction of a more distant relative of the true label.
Deng~\etal~\cite{deng2010does} simply train kNN- and SVM-based classifiers to minimise the expected WordNet LCA height directly.
Zhao~\etal~\cite{zhao2011large} modify standard multi-class logistic regression by replacing the output class probabilities with normalised class-similarity-weighted sums.
They also regularise feature selection using an ``overlapping-group-lasso penalty" which encourages the use of similar features for closely related classes, a rare example of a hierarchical regulariser.
Verma~\etal~\cite{verma2012learning}~incorporate normalised LCA height into a ``context-sensitive loss function" while learning a separate metric at each node in a taxonomy tree for nearest-neighbour classification.
Wu~\etal~\cite{wu2016learning} implement granular classification of food images by sharing a standard deep network backbone between multiple fully-connected layers, each one outputting class probabilities at its respective hierarchy level.
A separate label propagation step is used to smooth inconsistencies in the resulting marginal probabilities.
Alsallakh~\etal~\cite{alsallakh2017convolutional} likewise use a standard deep architecture as their starting point,
but instead add branches strategically to intermediate pipeline stages.
They thereby force the net to classify into offline-determined superclasses at the respective levels, backpropagating error in these intermediate predictions accordingly. 
At test time, these additions are simply discarded.

\subsection{Hierarchical architectures}
\label{subsec:divide_and_conquer}

These methods attempt to incorporate class hierarchy into the classifier architecture without necessarily changing the loss function otherwise.
The core idea is to ``divide and conquer" at the structural level, with the classifier assigning inputs to superclasses at earlier layers and making fine-grained distinctions at later ones.
In the context of language models, it was noted at least as early as Goodman~\cite{goodman2001classes} that classification with respect to an IS-A hierarchy tree could be formulated as a tree of classifiers outputting conditional probabilities, with the product of the conditionals along a given leaf's ancestry representing its posterior; motivated by efficiency, Morin \& Bengio~\cite{morin2005hierarchical} applied this observation to a binary hierarchy derived from WordNet.
Redmon \& Farhadi~\cite{redmon2017yolo9000} propose a modern deep-learning variant of this framework in the design of the YOLOv2 object detection and classification system.
Using a version of WordNet pruned into a tree,
they effectively train a conditional classifier at every parent node in the tree by using one softmax layer per sibling group and training under the usual cross-entropy loss over leaf posteriors.
While their main aim is to enable the integration of the COCO detection dataset with ImageNet, they suggest that graceful degradation on new or unknown object categories might be an incidental benefit.
Brust \& Denzler~\cite{brust2018integrating} propose an extension of conditional classifier chains to the more general case of DAGs.

The above approaches can be seen as a limiting case of hierarchical classification, in which every split in the hierarchy is cast as a separate classification problem.
Many hierarchical classifiers fall between this extreme and that of flat classification, working in terms of a coarser-grained conditionality in which a ``generalist" makes assignments to groupings of the target classes before then distinguishing the group members from one another using ``experts".
Xiao~\etal~\cite{xiao2014error}, the quasi-ensemble section of Hinton~\etal~\cite[Sec.~5]{hinton2015distilling},
Yan~\etal~\cite{yan2015hd}, and Ahmed~\etal~\cite{ahmed2016network} all represent modern variations on this theme (which first appears no later than~\cite{jacobs1991adaptive}).
Additionally, the listed methods all use some form of low-level feature sharing either via architectural constraint or parameter cloning, and all infer the visual hierarchy dynamically through confusion clustering or latent parameter inference.
Alsallakh~\etal~\cite{alsallakh2017convolutional} make the one proposal of which we are aware which combines hierarchical architectural modifications (at \emph{train} time) with a hierarchical loss, as described in Sec.~\ref{subsec:explicit_hierarchical_output_penalty}.
At test time, however, the architecture is that of an unmodified AlexNet, and all superclass ``assignment" is purely implicit.

\section{Method}
\label{sec:method}

We now outline two simple methods that allow
to leverage class hierarchies in order to make better mistakes on image classification.
We concentrate on the case where the output of the network is a categorical distribution over classes for each input image and denote the corresponding distribution as $p(C) = \phi_C(x; \theta)$, where subscripts denote vector indices and $x$ and $\theta$ are omitted.
In Sec.~\ref{sec:hxe}, we describe the \emph{hierarchical cross-entropy} (HXE), a straightforward example of the hierarchical losses reviewed in Sec.~\ref{subsec:explicit_hierarchical_output_penalty}.
This approach expands each class probability into the chain of conditional probabilities defined by its lineage in a given hierarchy tree.
It then reweights the corresponding terms in the loss so as to penalise classification mistakes in a way that is informed by the hierarchy.
In Sec.~\ref{sec:soft}, we suggest an easy choice of embedding function to implement the label-embedding framework of Sec.~\ref{subsec:soft_targets_embeddings}.
The resulting \emph{soft labels} are PMFs over $\mathcal C$ whose values decay exponentially w.r.t.~an LCA-based distance to the ground truth.

\subsection{Hierarchical cross-entropy}
\label{sec:hxe}

When the hierarchy $\mathcal H$ is a tree, it corresponds to a unique factorisation of the categorical distribution $p(C)$ over classes in terms of the conditional probabilities along the path connecting each class to the root of the tree.
Denoting the path from a leaf node $C$ to the root $R$ as $C^{(0)} = C, \ldots, C^{(h)} = R$, the probability of class $C$ can be factorised as
\begin{equation} \label{eq:factorisation}
	p(C) = \prod_{l=0}^{h-1} p (C^{(l)}| C^{(l+1)}),
\end{equation}
where $h \equiv h(C)$ is the depth of node $C$.
Note that we have omitted the last term $p(C^{(h)}) = 1$.
The conditionals can conversely be written in terms of the class probabilities as
\begin{equation} \label{eq:conditionals}
	p (C^{(l)}|C^{(l+1)} ) = \frac{\sum_{A \in \text{Leaves}(C^{(l)})} p (A)} {\sum_{B \in \text{Leaves}(C^{(l+1)})} p (B)},
\end{equation}
where $\text{Leaves}(C)$ denotes the set of leaf nodes of the subtree starting at node $C$.

A direct way to incorporate hierarchical information in the loss is to hierarchically factorise the output of the classifier according to Eqn.~\ref{eq:factorisation} and define the total loss as the reweighted sum of the cross-entropies of the conditional probabilities.
This leads us to define the \emph{hierarchical cross-entropy} (HXE) as 
\begin{equation}  \label{eq:HXE}
	\mathcal L_{\text{HXE}}(p, C) = -\sum_{l=0}^{h-1} \lambda(C^{(l)}) \log p (C^{(l)}|C^{(l+1)}),
\end{equation}
where $ \lambda(C^{(l)})$ is the weight associated with the edge node $C^{(l+1)} \to C^{(l)}$, see Fig.~\ref{fig:losses}\red{a}.
Although this loss is expressed in terms of conditional probabilities, it can easily be applied to models that output class probabilities using Eqn.~\ref{eq:conditionals}.
Note that $\mathcal L_{\text{HXE}}$ reduces to the standard cross-entropy when all weights are equal to \num{1}.
This limit case, which was briefly mentioned by Redmon \& Farhadi in their \yolo{} paper~\cite{redmon2017yolo9000}, results only in architectural changes but does not incorporate hierarchical information in the loss directly.

Eqn.~\ref{eq:HXE} has an interesting information-theoretical interpretation: since each term $\log p (C^{(l)} |C^{(l+1)} )$ corresponds to the the information associated with the edge $C^{(l+1)} \to C^{(l)}$ in the hierarchy, the HXE corresponds to discounting the information associated with each of these edges differently.
Note that since the HXE is expressed in terms of conditional probabilities, the reweighting in Eqn.~\ref{eq:HXE} is not equivalent to reweighting the cross-entropy for each possible ground truth class independently (as done, for instance, in~\cite{lin2017focal,cui2019class}).
A sensible choice for the weights is to take
\begin{equation} \label{eq:lambda}
	\lambda(C) = \exp(- \alpha h(C)),
\end{equation}
where $h(C)$ is the depth of node $C$ and $\alpha>0$ is a hyperparameter that controls the extent to which information is discounted down the hierarchy.
The higher the value of $\alpha$, the higher the preference for ``generic'' as opposed to ``fine-grained'' information, because classification errors related to nodes further away from the root receive a lower loss.
While such a definition has the advantages of interpretability and simplicity, one could think of other meaningful weightings (\eg based on the branching factor of the hierarchy tree).
We concentrate on Eqn.~\ref{eq:lambda} here, while leaving the exploration of different strategies for future work.

\subsection{Soft labels}
\label{sec:soft}

\begin{figure}[t]
	\centering
	\raisebox{2.8cm}{(a)}\hspace{-4mm}
	\begin{forest}
		[R[D, edge={line width=1.5}, edge label={node[midway,left,font=\scriptsize]{$\lambda(D)$}}[\scalebox{1.4}{$\substack{\underline{\text{A}}\\\shortparallel\\1}$\hspace{3mm}}, edge={line width=1.5}, edge label={node[midway,left,font=\scriptsize]{$\lambda(A)$}}][\scalebox{1.4}{$\substack{\text{B}\\\shortparallel\\0}$\hspace{3mm}}, edge label={node[midway,right,font=\scriptsize]{$\lambda(B)$}}]][\scalebox{1.4}{$\substack{\text{C}\\\shortparallel\\0}$\hspace{2mm}}, edge label={node[midway,right,font=\scriptsize]{$\lambda(C)$}}, l*=2.25]]
	\end{forest} \hfill
	\raisebox{2.8cm}{(b)}\hspace{-5mm}
	\begin{forest}
		[R[D, edge={line width=1.5}[\scalebox{1.4}{$\substack{\underline{\text{A}}\\\shortparallel\\\scriptscriptstyle{y^{\text{soft}}_A(A)}}$}, edge={line width=1.5}][\scalebox{1.4}{$\substack{\text{B}\\\shortparallel\\\scriptscriptstyle{y^{\text{soft}}_B(A)}}$}, edge={line width=1.5}]][\scalebox{1.4}{$\substack{\text{C}\\\shortparallel\\ \scriptscriptstyle{y^{\text{soft}}_C(A)}}$}, edge={line width=1.5}, l*=2.35]]
	\end{forest}
	\caption{\label{fig:losses}
		Representations of the \emph{HXE} (Sec.~\ref{sec:hxe}) and \emph{soft labels} (Sec.~\ref{sec:soft}) losses for a simple illustrative hierarchy are drawn in subfigures (a) and (b) respectively.
		The ground-truth class is underlined, and the edges contributing to the total value of the loss are drawn in bold.}
\end{figure}
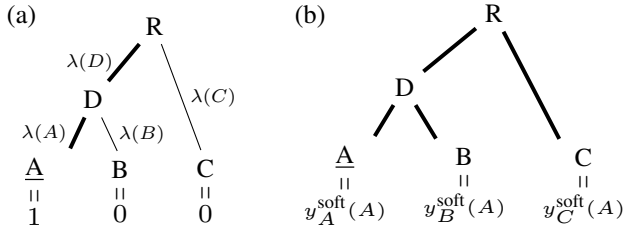

Our second approach to incorporating hierarchical information, \emph{soft labels}, is a label-embedding approach as described in Sec.~\ref{subsec:soft_targets_embeddings}.
These methods use a mapping function $y(C)$ to associate classes with representations which encode class-relationship information that is absent in the trivial case of the one-hot representation.
In the interest of simplicity, we choose a mapping function $y^{\text{soft}}(C)$ which outputs a categorical distribution over the classes.
This enables us to simply use the standard cross-entropy loss:
\begin{equation} \label{eq:soft_x_ent}
	\mathcal L_{\text{Soft}}(p, C) = -\sum_{A \in \mathcal C} y^{\text{soft}}_A (C) \log p(A),
\end{equation}
where the soft label embedding is given componentwise by
\begin{equation} \label{eq:softtarget}
	y^{\text{soft}}_A(C) = \frac{\exp( - \beta d(A, C))}{\sum_{B\in \mathcal C} \exp(-\beta d(B, C))},
\end{equation}
for class distance function $d$ and parameter $\beta$.
This loss is illustrated in Fig.~\ref{fig:losses}\red{b}.
For the distance function $d(C_i, C_j)$, we use the height of LCA$(C_i, C_j)$ divided by the height of the tree.
To understand the role of the hyperparameter $\beta$, note that values of $\beta$ that are much bigger than the typical inverse distance in the tree result in a label distribution that is nearly one-hot,~\ie $y_A(C) \simeq \delta_{AC}$, in which case the cross-entropy reduces to the familiar single-term log-loss expression.
Conversely, for very small values of $\beta$ the label distribution is near-uniform.
Between these extremes, greater probability mass is assigned to classes more closely related to the ground truth, with the magnitude of the difference controlled by $\beta$.

We offer two complementary interpretations that motivate this representation (besides its ease).
For one, the distribution describing each target class can be considered to be a model of the actual uncertainty that a labeller would experience due to visual confusion between closely related classes\footnote{In a recent work, Peterson \etal~\cite{peterson2019human} make use of soft labels expressing the distribution of human labellers for a subset of CIFAR-10, showing strong generalisation for classifiers trained on them.}.
It could also be thought of as encoding the extent to which a common response to different classes is \emph{required} of the classifier,~\ie~the imposition of correlations between outputs, where higher correlations are expected for more closely related classes.
This in turn suggests a connection to the superficially different but conceptually related \emph{distillation} method of Hinton~\etal~\cite[Sec.~2]{hinton2015distilling}, in which correlations between a large network's responses to different classes are mimicked by a smaller network to desirable effect.
Here, we simply supply these correlations directly, using widely available hierarchies. 

Another important connection is the one to \emph{label smoothing}~\cite{szegedy2016rethinking}, in which one-hot labels are combined with the uniform distribution.
This technique has been used to regularise large neural networks (\eg~\cite{szegedy2016rethinking,chorowski2017towards,vaswani2017attention,zoph2018learning}), but has only recently~\cite{muller2019does} been studied more thoroughly.

\section{Evaluation}
\label{sec:experiments}

In the following, we first describe the datasets (Sec.~\ref{sec:dataset}) and metrics (Sec.~\ref{sec:metrics}) of the setup common to all of our experiments.
Then, in Sec.~\ref{sec:performance}, we evaluate our two simple proposals and compare them to the prior art.
Finally, we experiment with random hierarchies to understand when information on class relatedness can help classification.

\subsection{Datasets}
\label{sec:dataset}

In our experiments, we use tieredImageNet~\cite{ren2018meta} (a large subset of ImageNet) and iNaturalist'19~\cite{van2018inaturalist}, two datasets with hierarchies that are \emph{a)} significantly different from one another and \emph{b)} complex enough to cover a large number of visual concepts.
ImageNet aims to populate the WordNet~\cite{miller1998wordnet} hierarchy of nouns, with WordNet itself generated by inspecting IS-A lexical relationships.
By contrast, iNaturalist'19~\cite{van2018inaturalist} has a biological taxonomy~\cite{ruggiero2015higher} at its core.

\smallpar{tieredImageNet}
was originally
introduced by Ren~\etal~\cite{ren2018meta} for the problem of few-shot classification, in which the sets of classes between dataset splits are disjoint.
The authors' motivation in creating the dataset was to use the WordNet hierarchy  to generate splits containing significantly different classes, facilitating better assessment of few-shot classifiers by enforcing problem difficulty.

Although our task and motivations are different, we chose this dataset because of the large portion of the WordNet hierarchy spanned by its classes.
To make it suitable for the problem of (standard) image classification, we re-sampled the dataset so as to represent all classes across the train, validation, and test splits.
Moreover, since the methods proposed in Sec.~\ref{sec:hxe} and \yolo~\cite{redmon2017yolo9000} require that the graph representing the hierarchy be a tree, we modified the graph of the spanned WordNet hierarchy slightly to comply with this assumption (more details available in Appendix~\ref{sec:prune_wordnet}).
After this procedure, we obtained a tree of height \num{13} covering \num{608} classes.
We refer to this modified version of tieredImageNet as \tih.

\smallpar{iNaturalist} is a dataset of images of organisms that has mainly been used to evaluate fine-grained visual categorisation methods.
The dataset construction protocol differs significantly from the one used for ImageNet in that it relies 
on passionate volunteers instead of workers paid per task~\cite{van2018inaturalist}.
Importantly, for the 2019 edition of the CVPR Fine-Grained Visual Categorization Workshop, metadata with hierarchical relationships between species have been released.
In contrast to WordNet, this taxonomy is an 8-level complete tree spanning \num{1010} classes that can readily be used in our experiments without modifications.
Since the labels for the test set are not public, we randomly re-sampled three splits from the original train and validation splits into a new training, validation and test set (with respective probabilities of \num{0.7}, \num{0.15}, and \num{0.15}) 

\subsection{Metrics}
\label{sec:metrics}

We consider three measures of performance, covering different interpretations of a classifier's \emph{mistakes}.

\smallpar{Top-$k$ error.}
Under this measure, an example is defined as correctly classified if the ground truth is among the top $k$ classes with the highest likelihood.
This is the measure normally used to compare classifiers, usually with $k{=}1$ or $k{=}5$.
Note that this measure considers all mistakes of the classifier equally, irrespective of how ``similar'' the predicted class is to the ground truth.

\smallpar{Hierarchical measures.}
We also consider measures that, in contrast to the top-$k$ error, do weight the severity of mistakes.
We use the height of the lowest common ancestor (LCA) between the predicted class and the ground truth as a core severity measure, as originally proposed in the papers describing the creation of ImageNet~\cite{deng2009imagenet,deng2010does}.
As remarked in~\cite{deng2010does}, this measure should be thought of in logarithmic terms, as the number of confounded classes is exponential in the height of the ancestor.
We also experimented with the Jiang-Conrath distance as suggested by Deselaers \& Ferrari~\cite{deselaers2011visual}, but did not observe meaningful differences wrt. the height of the LCA.

We consider two measures that utilise the height of the LCA between nodes in the hierarchy.
\begin{tight_it}
		\item The \textbf{hierarchical distance of a mistake} is the mean height of the LCA between the ground truth and the predicted class \emph{when the input is misclassified}, \ie when the class with the maximum likelihood is incorrect.
		Hence, it measures the severity of misclassification when only a single class can be considered as a prediction.
		\item The \textbf{average hierarchical distance of top-$k$}, instead, takes the mean LCA height between the ground truth and each of the $k$ most likely classes.
		This measure can be important when multiple hypotheses of a classifier can be considered for a certain downstream task.
\end{tight_it}

\subsection{Experimental results}
\label{sec:performance}

In the following, we analyse the performance of the two approaches described in Sec.~\ref{sec:hxe} and Sec.~\ref{sec:soft}, which we denote by \textit{HXE} and \textit{soft labels}, respectively.
Besides a vanilla cross-entropy-based flat classifier, we also implemented and compared against the methods proposed by Redmon \& Farhadi~\cite{redmon2017yolo9000} (\yolo)\footnote{Note that this refers to the conditional classifier subsystem proposed in Sec. 4 of that work, not the main object detection system.}, Frome~\etal~\cite{frome2013devise} (\devise), and Barz \& Denzler~\cite{barz2019hierarchy}.
As mentioned in Sec.~\ref{sec:introduction}, these methods represent, to the best of our knowledge, the only modern attempts to deliberately reduce the semantic severity of a classifier's mistakes that are generally applicable to any modern architecture.
Note, though, that we do not run \devise{} on \inh{}, as the class IDs of this dataset are alien to the corpus used by word2vec~\cite{mikolov2013efficient}.

Finally, we do not compare against the ``generalist/expert" architectures surveyed in Sec.~\ref{subsec:divide_and_conquer} for reasons explained in Appendix~\ref{sec:note_hier_arch}.

Since we are interested in understanding  the mechanisms by which the above metrics can be improved, it is essential to use a simple configuration that is common between all of the algorithms taken into account.
We use a ResNet-18 architecture (with weights pretrained on ImageNet) trained with Adam~\cite{reddi2019convergence} for \num[group-separator={,}]{200000} steps and mini-batches of size \num{256}.
We use a learning rate of $1\mathrm{e}{-5}$ unless specified otherwise.
Further implementation details are deferred to Appendix~\ref{sec:more_details}.

\begin{figure}[h]
		\centering
		\includegraphics[width=0.33\textwidth]{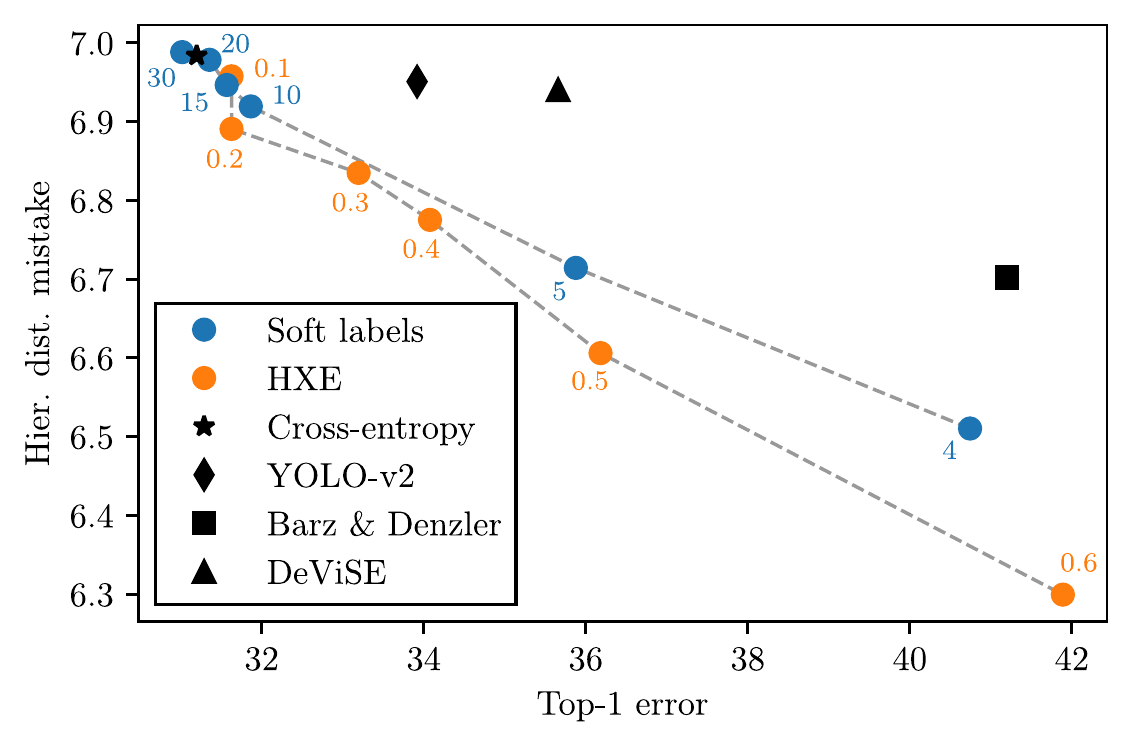} \\
		\includegraphics[width=0.33\textwidth]{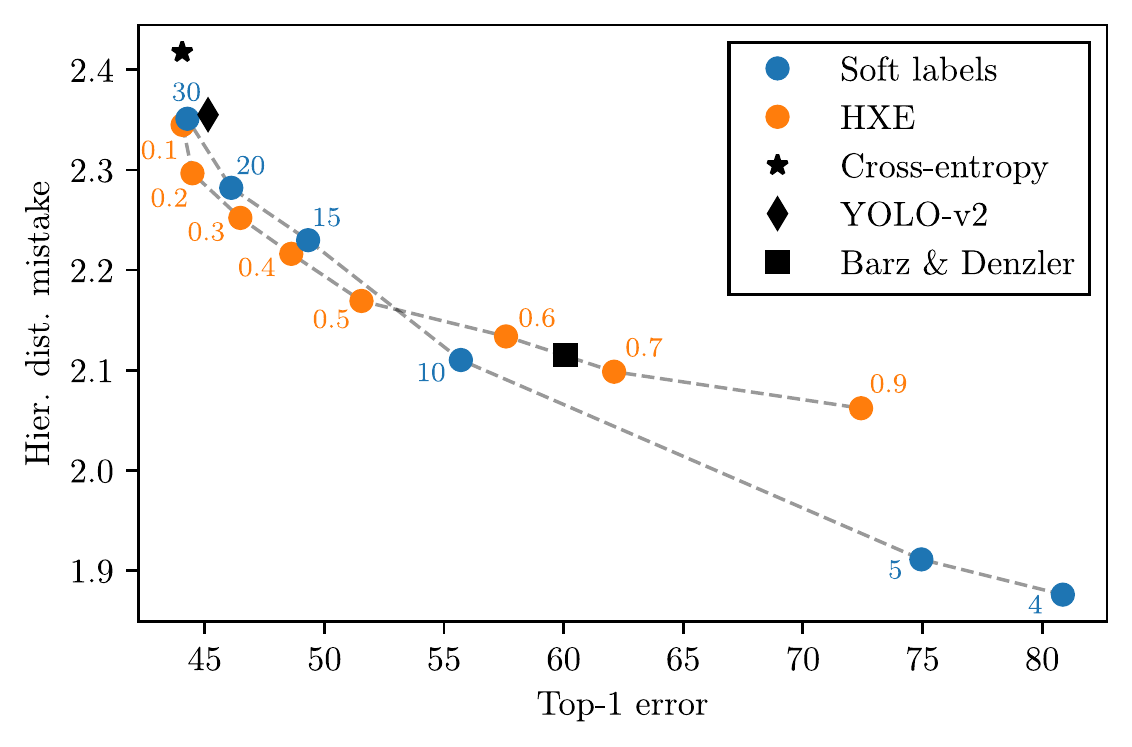}
		\vspace{-0.3cm}
		\caption{\label{fig:tradeoff-mistakes}
			Top-1 error vs. hierarchical distance of mistakes, for \tih{} (top) and \inh{} (bottom).
		Points closer to the bottom-left corner of the plot are the ones achieving the best tradeoff.
		}
\end{figure}

\begin{figure*}[h]
        \centering
		\includegraphics[width=0.33\textwidth]{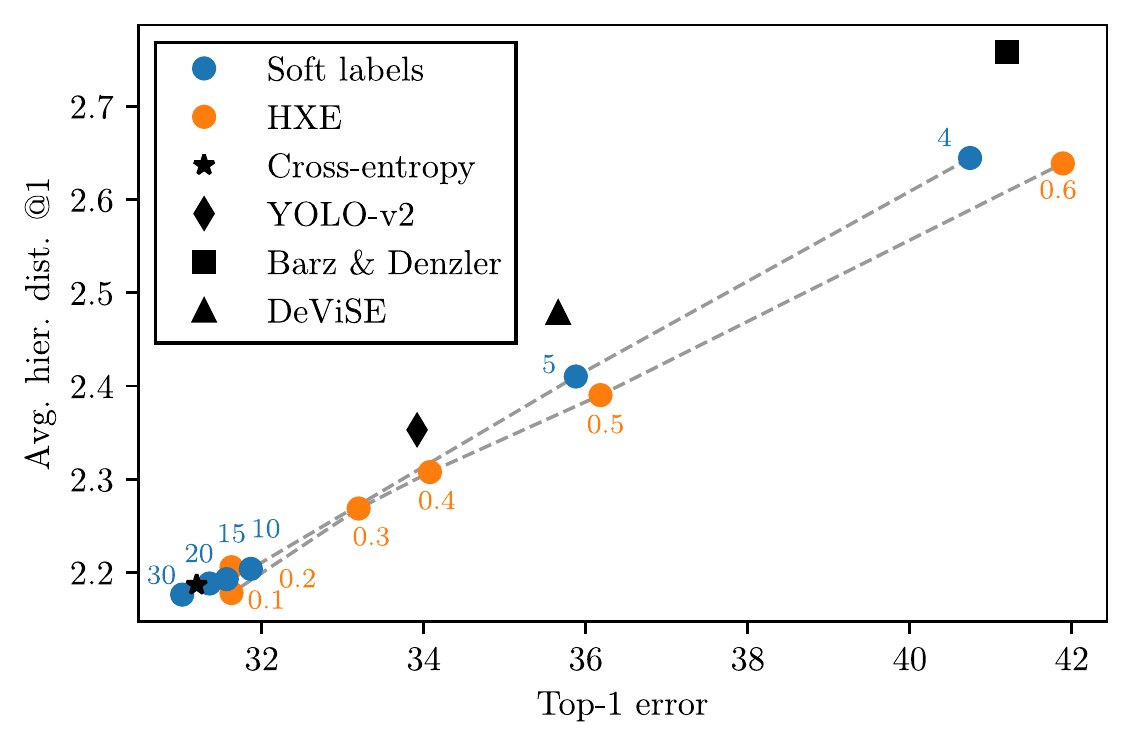}
		\includegraphics[width=0.33\textwidth]{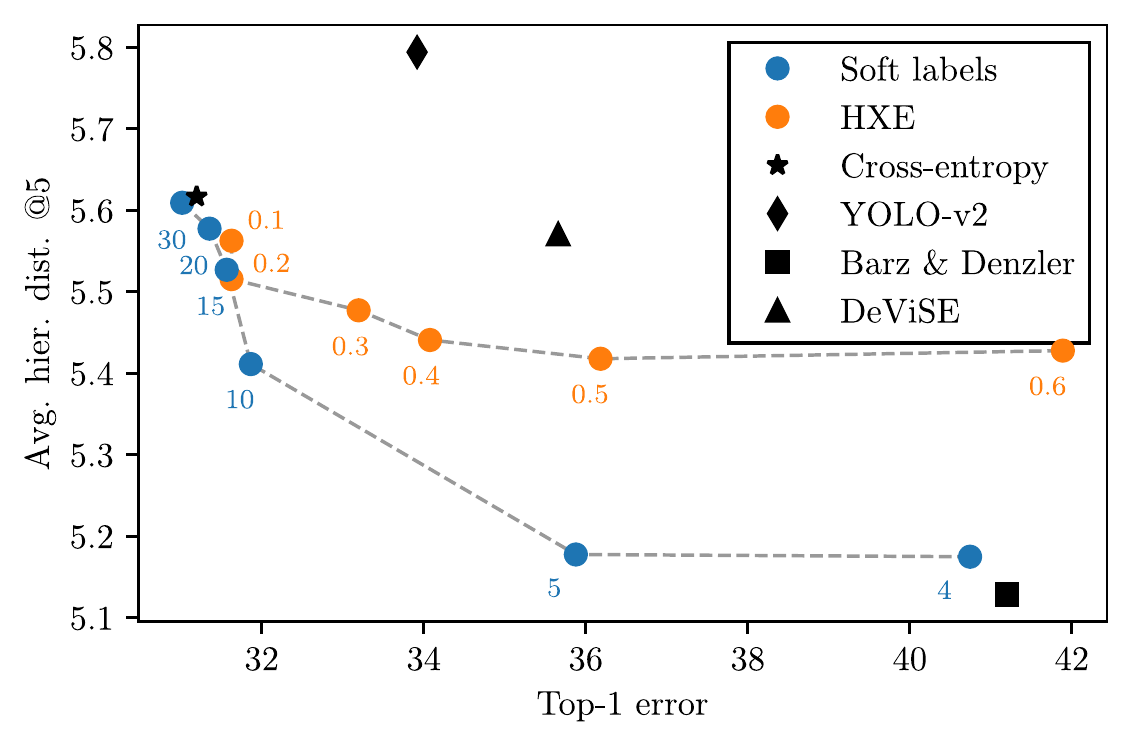}
		\includegraphics[width=0.33\textwidth]{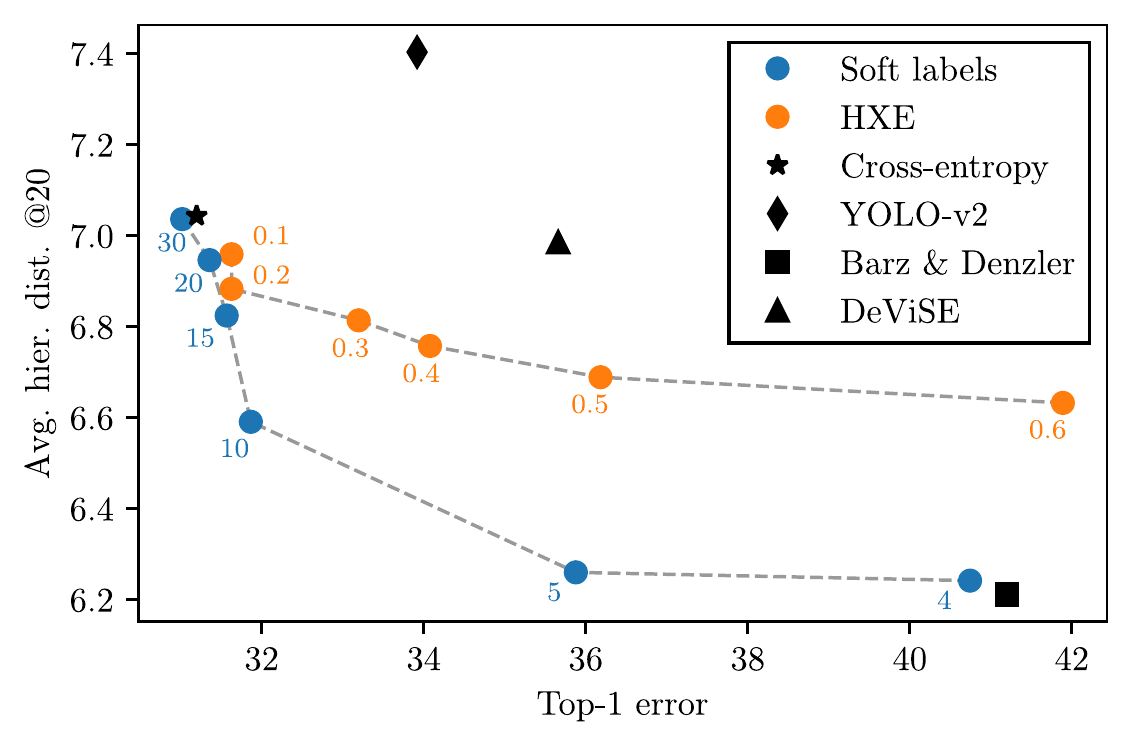}
		\includegraphics[width=0.33\textwidth]{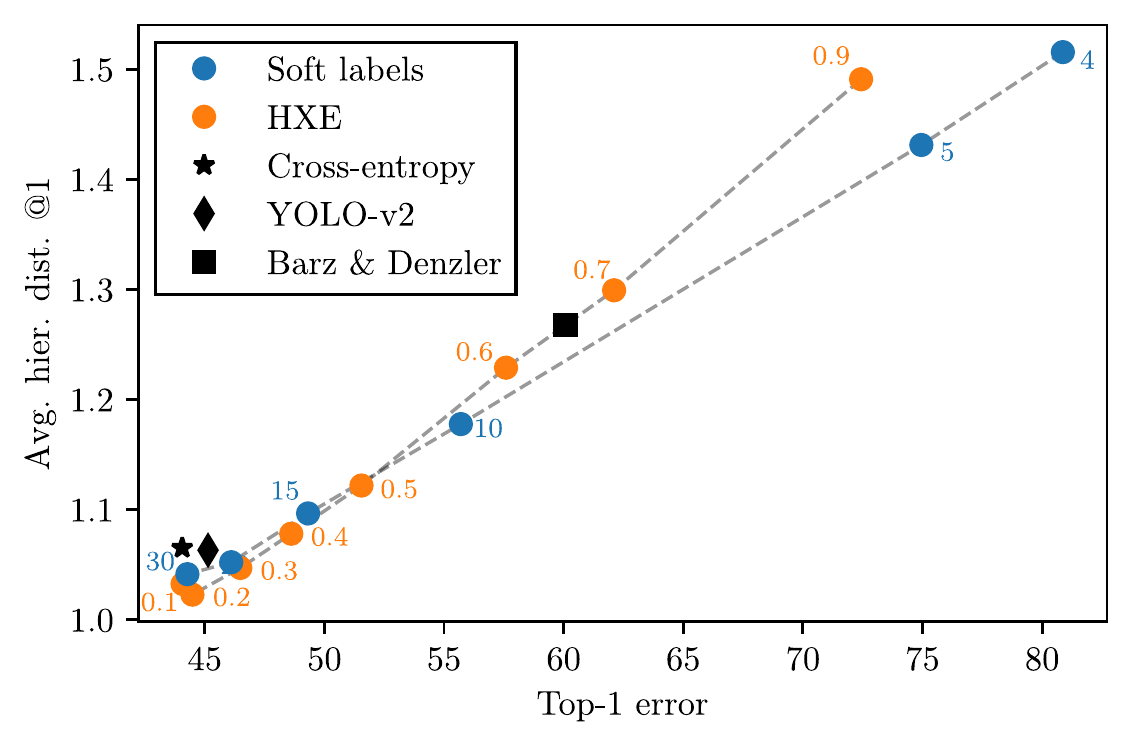}
		\includegraphics[width=0.33\textwidth]{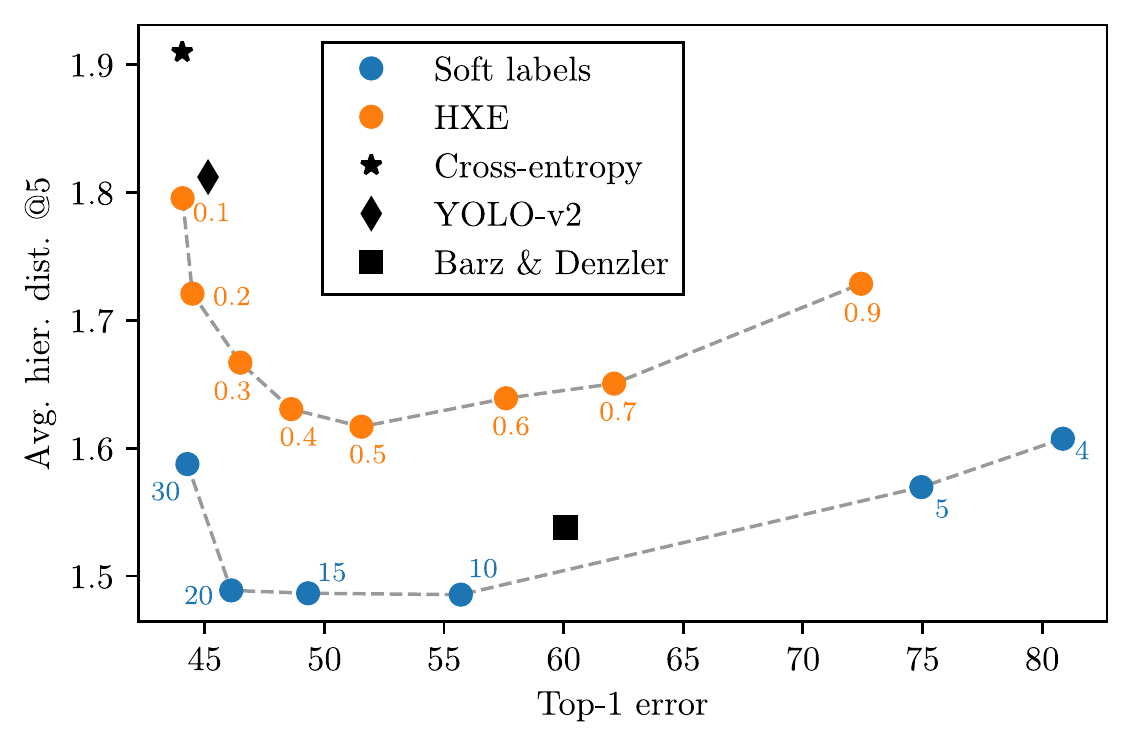}
		\includegraphics[width=0.33\textwidth]{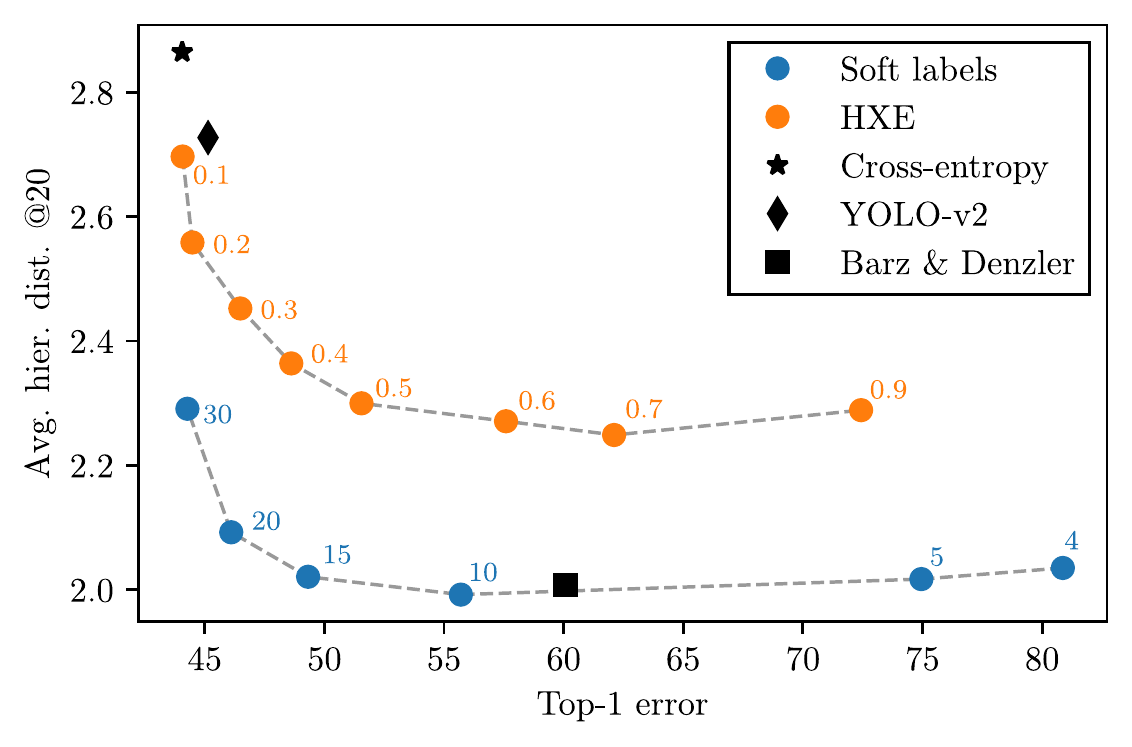}
		\vspace{-0.5cm}
		\caption{\label{fig:tradeoff-dist}
			Top-1 error vs. average hierarchical distance of top-$k$ (with $k \in \{1, 5, 20\}$)  for \tih{} (top three) and \inh{} (bottom three).
			Points closer to the bottom-left corner of the plot are the ones achieving the best tradeoff.
		}
\end{figure*}

\smallpar{Main results.}
In Fig.~\ref{fig:tradeoff-mistakes} and~\ref{fig:tradeoff-dist} we show how it is possible to effectively trade off top-1 error to reduce hierarchical error, by simply adjusting the hyperparameters $\alpha$ and $\beta$ in Eqn.~\ref{eq:lambda} and~\ref{eq:softtarget}.
Specifically, increasing $\alpha$ corresponds to (exponentially) discounting information down the hierarchy, thus more severely penalising mistakes where the predicted class is further away from the ground truth.
Similarly, decreasing $\beta$ in the soft-label method amounts to progressively shifting the label mass away from the ground truth and towards the neighbouring classes.
Both methods reduce to the cross-entropy in the respective limits $\alpha \to 0$ and $\beta \to \infty$.
Moreover, notice that varying $\beta$ affects the entropy of the distribution representing a soft label, where the two limit cases are $\beta{=}\infty$ for the standard one-hot case and $\beta{=}0$ for the uniform distribution.
We experiment with $0.1\le\alpha\le0.9$ and $4\le\beta\le30$.

To limit noise in the evaluation procedure, for both of our methods and all of the competitors, we fit a 4th-degree polynomial to the validation loss (after having discarded the first \num[group-separator={,}]{50000} training steps) and pick the epoch corresponding to its minimum along with its four neighbours.
Then, to produce the points reported in our plots, we average the results obtained from these five epochs on the validation set, while reserving the test set for the experiments of Table~\ref{tab:tiered_inat}.
Notice how, in Fig.~\ref{fig:tradeoff-dist}, when considering the hierarchical distance with $k{=}1$, methods are almost perfectly aligned along the plot diagonal, which demonstrates the strong linear correlation between this metric and the top-1 error.
This result is consistent with what is observed in~\cite{russakovsky2015imagenet}, which in \num{2011} led the organisers of the ILSVRC workshop to discard rankings based on hierarchical distance.

When considering the other metrics described in Sec.~\ref{sec:metrics}, a different picture emerges.
In fact, a tradeoff between top-1 error and hierarchical distance is evident in Fig.~\ref{fig:tradeoff-mistakes} and in the plots of Fig.~\ref{fig:tradeoff-dist} with $k{=}5$ and $k{=}20$.
Notice how the points on the plots belonging to our methods outline a set of tradeoffs that subsumes the prior art.
For example, in Fig.~\ref{fig:tradeoff-mistakes}, given any desired tradeoff betweeen top-1 error and hierarchical distance of mistakes on \tih, it is better to use HXE than any other method.
A similar phenomenon is observable when considering the average hierarchical distance of top-5 and top-20 (Fig.~\ref{fig:tradeoff-dist}), although in these cases it is better to use the \emph{soft labels}.
The only exception to this trend is represented by Barz \& Denzler~\cite{barz2019hierarchy} on \tih, which can achieve slightly lower average hierarchical distance for $k{=}5$ or $k{=}20$ than \emph{soft labels} with $\beta{=}5$ at a significant cost in terms of top-1 error.

\begin{table*}[tb]
\caption{Results on the test sets of \tih{} (top) and \inh{} (bottom), with $95\%$ confidence intervals.
For each column of each dataset, the best entry is hightlighted in \hlyellow{yellow}, while the worst is highlighted in \hlcyan{gray}.}
\label{tab:tiered_inat}
\vspace{-0.08in}
\begin{center}
\begin{scriptsize}
\vspace{-0.3cm}
\begin{tabular}{l||c||ccc||c}
\toprule
																			  & Hier. dist. mistake \DA & Avg. hier. dist. @1 \DA & Avg. hier. dist. @5  \DA & Avg. hier. dist. @20  \DA & Top-1 error \DA \\
\midrule
		\textbf{\textsc{Cross-entropy}}                                               & $6.97\pm0.005$         & \cellcolor{yellow}$ 2.20\pm0.001$         & $ 5.63\pm0.004$          & $ 7.05\pm0.011$           & $\cellcolor{yellow}31.58\pm0.137$ \\

		\textbf{\textsc{\BD}~\cite{barz2019hierarchy}}                        & $ 6.69\pm0.028$         & \cellcolor{lightgray}$ 2.78\pm0.008$         & \cellcolor{yellow} $ 5.13\pm0.007$          & $ 6.21\pm0.006$           &\cellcolor{lightgray} $41.53\pm0.207$ \\

		\textbf{\textsc{\yolo}~\cite{redmon2017yolo9000}}                          & $ 6.94\pm0.009$         & $ 2.38\pm0.007$         & \cellcolor{lightgray}$ 5.82\pm0.019$          & $\cellcolor{lightgray}7.40\pm0.025$           & $34.24\pm0.009$\\

\textbf{\textsc{\devise}~\cite{frome2013devise}}                           & $ 6.92\pm0.011$         & $ 2.51\pm0.011$         & $ 5.58\pm0.002$          & $ 6.99\pm0.001$           & $36.27\pm0.130$\\

\textbf{\textsc{HXE $\alpha{=}0.2$}} (ours)                                  & $ 6.87\pm0.018$         & $ 2.22\pm0.010$         & $ 5.53\pm0.004$          & $ 6.89\pm0.005$           & $32.31\pm0.160$ \\

		\textbf{\textsc{HXE $\alpha{=}0.5$}} (ours)                           & $\cellcolor{yellow}6.58\pm0.036$         & $ 2.45\pm0.020$         & $ 5.44\pm0.003$          & \cellcolor{yellow}$ 6.14\pm0.008$           & $37.24\pm0.490$ \\

		\textbf{\textsc{Soft-labels $\beta{=}30$}} (ours)                            & \cellcolor{lightgray} $ 6.98\pm0.013$         & $ 2.21\pm0.004$         & $ 5.62\pm0.007$          & $ 7.04\pm0.008$           & $31.72\pm0.046$ \\

\textbf{\textsc{Soft-labels $\beta{=}5$}} (ours)                             & $ 6.62\pm0.008$         & $ 2.47\pm0.005$         & $ 5.19\pm0.007$          & $ 6.26\pm0.004$           & $37.29\pm0.064$ \\
\midrule\midrule
		\textbf{\textsc{Cross-entropy}}                                               & \cellcolor{lightgray}$2.43\pm0.003$         &$ 1.07\pm0.006$         &\cellcolor{lightgray} $ 1.91\pm0.004$          &\cellcolor{lightgray} $ 2.87\pm0.007$           & \cellcolor{yellow} $44.04\pm0.197$ \\

		\textbf{\textsc{\BD}~\cite{barz2019hierarchy}}                        & $ 2.11\pm0.008$         & \cellcolor{lightgray}$ 1.27\pm0.006$         & $ 1.54\pm0.005$          & $ 2.01\pm0.003$           &\cellcolor{lightgray} $60.11\pm0.188$ \\

\textbf{\textsc{\yolo}~\cite{redmon2017yolo9000}}                          & $ 2.36\pm0.006$         & $ 1.06\pm0.004$         & $ 1.81\pm0.007$          & $ 2.73\pm0.012$           & $45.00\pm0.152$ \\

		\textbf{\textsc{HXE} $\alpha{=}0.1$} (ours)                                   & $ 2.33\pm0.007$         &\cellcolor{yellow} $ 1.03\pm0.008$         & $ 1.79\pm0.004$          & $ 2.70\pm0.004$           & \cellcolor{yellow} $44.01\pm0.241$ \\

\textbf{\textsc{HXE} $\alpha{=}0.5$} (ours)                                   & $ 2.16\pm0.001$         & $ 1.13\pm0.005$         & $ 1.81\pm0.004$          & $ 2.30\pm0.003$           & $52.06\pm0.225$ \\

		\textbf{\textsc{Soft-labels} $\beta{=}30$} (ours)                            & $ 2.35\pm0.005$         & \cellcolor{yellow}$1.03\pm0.004$         & $ 1.58\pm0.004$          & $ 2.29\pm0.003$           & \cellcolor{yellow} $44.00\pm0.084$ \\

		\textbf{\textsc{Soft-labels} $\beta{=}10$} (ours)                            & \cellcolor{yellow}$ 2.10\pm0.007$         & $ 1.16\pm0.007$         & \cellcolor{yellow}$ 1.48\pm0.005$          & \cellcolor{yellow} $ 1.99\pm0.005$           & $55.54\pm0.173$ \\

\bottomrule
\end{tabular}

\end{scriptsize}
\end{center}
\vspace{-0.08in}
\end{table*}


Using the results illustrated in Fig.~\ref{fig:tradeoff-mistakes} and~\ref{fig:tradeoff-dist}, we pick two reasonable operating points for both of our proposals: one for the high-distance/low-top1-error regime, and one for the low-distance/high-top1-error regime. 
We then run both of these configurations on the test sets and report our results in Table~\ref{tab:tiered_inat}.
Means and 95\% confidence intervals are obtained from the five best epochs.

The trends observed on the validation set largely repeat themselves on the test set.
When one desires to prioritise top-1 error, then soft labels with high $\beta$ or HXE with low $\alpha$ are more appropriate, as they outperform the cross-entropy on the hierarchical-distance-based metrics while being practically equivalent in terms of top-1 error. 
In cases where the hierarchical measures should be prioritised instead, it is preferable to use soft labels with low $\beta$ or HXE with high $\alpha$, depending on the particular choice of hierarchical metric.
Although the method of Barz \& Denzler is competitive in this regime, it also exhibits the worst deterioration in top-1 error with respect to the cross-entropy.

Our experiments generally indicate, over all tested methods, an inherent tension between performance in the top-1 sense and in the hierarchical sense.
We speculate that there may be a connection between this tension and observations proceeding from the study of adversarial examples indicating a tradeoff between robustness and (conventional) accuracy, as in~\eg~\cite{tsipras2018robustness,zhang2019theoretically}.

\smallpar{Can hierarchies be arbitrary?}
Although the lexical WordNet hierarchy and the biological taxonomy of iNaturalist are not visual hierarchies per se, they reflect visual relationships between the objects represented in the underlying datasets.
Since deep networks leverage visual features, it is interesting to investigate the extent to which the structure of a particular hierarchy is important.
The connection between visual and semantic proximity has also been explored in such works as \cite{deng2010does,deselaers2011visual}.
But what happens if we impose an arbitrary hierarchy that potentially subverts this relationship?

\begin{figure}[t]
		\centering
		\includegraphics[width=0.33\textwidth]{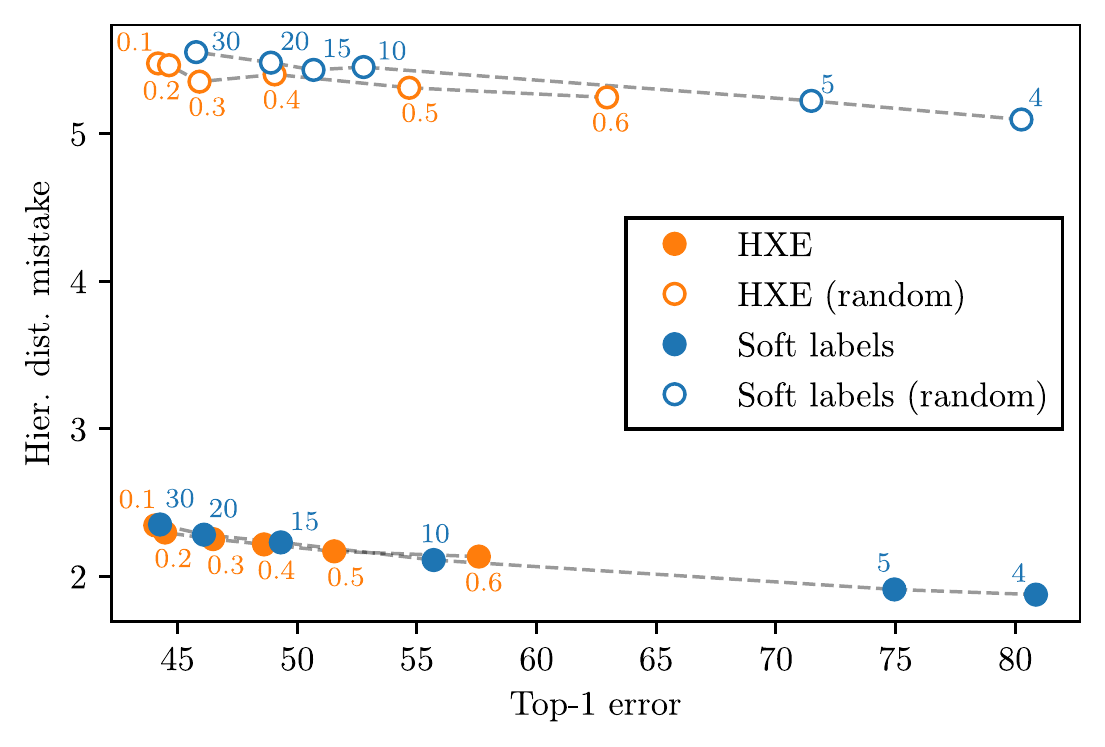}
        \includegraphics[width=0.33\textwidth]{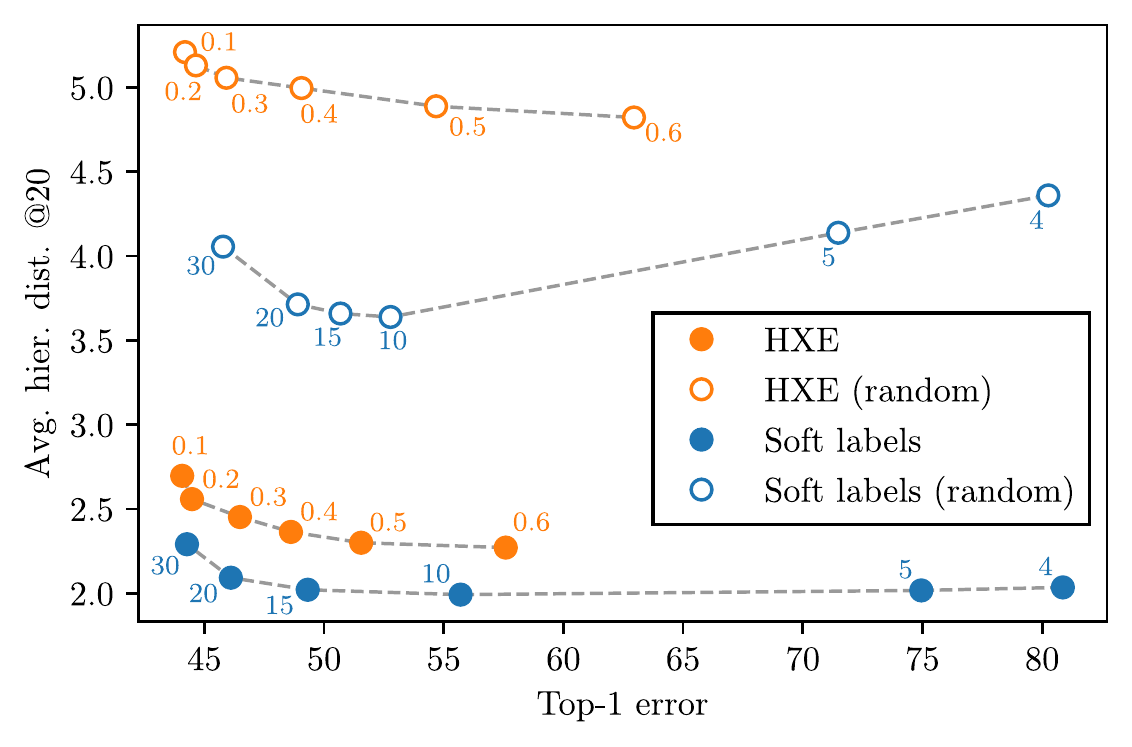}
        \vspace{-0.3cm}
		\caption{Top-1 error vs. hierarchical distance of mistakes (top) and hierarchical distance of top-20 (bottom) for \inh{}. Points closer to the bottom-left corner of the plots are the ones achieving the best tradeoff.}
		\label{fig:random_hier}
\end{figure}

To answer this question, we randomised the nodes of the hierarchies and repeated our experiments.
Results on \inh{} are displayed in Fig.~\ref{fig:random_hier} (\tih{} exhibits a similar trend).
Again, we report tradeoff plots showing top-1 errors on the x-axis and metrics based on the height of the LCA (on the randomised hierarchy) on the y-axis.
It is evident that the hierarchical distance metrics are significantly worse when using the random hierarchy.
Although this is not surprising, the extent to which the results deteriorate is remarkable.
This suggests that the inherent nature of the structural relationship expressed by a hierarchy is paramount for learning classifiers that, besides achieving competitive top-1 accuracy, are also able to make better mistakes.
Thus, while one may wish to enforce application-specific relationships using this approach (as motivated in Sec.~\ref{sec:introduction}), the effectiveness of doing so may be constrained by underlying properties of the data.

Curiously, for several values of the $\alpha$ and $\beta$ hyperparameters of HXE and \emph{soft labels}, the top-1 error of the random hierarchy is consistently \emph{lower} than its ``real'' hierarchy counterpart.
We speculate this might be due to the structural constraints imposed by a hierarchy anchored to the visual world, which can limit a neural network from opportunistically learning correlations that allow it to achieve low top-1 error (at the expense of ever more brittle generalisation).
Indeed, the authors of~\cite{zhang2016understanding} noted that it is more difficult to train a deep network to map real images to random labels than it is to do so with random images.
The most likely explanation for this is that common visual features, which are inescapably shared by closely related examples, dictate common responses.

\section{Conclusion}
\label{sec:conclusions}

Since the advent of deep learning, the community's interest in making better classification mistakes seems to have nearly vanished.
In this paper, we have shown that this problem is still very much open and ripe for a comeback.
We have demonstrated that two simple baselines that modify the cross-entropy loss are able to outperform the few modern methods tackling this problem.
Improvements in this task are undoubtedly possible, but it is important to note the delicate balance between standard top-1 accuracy and mistake severity.
As it stands, it appears that it is possible to make better mistakes, but the nature of the class relationships defining the concept of ``better" is crucial.
Our hope is that the results presented in this paper are soon to be surpassed by the new competitors that it has inspired.

{\small
\bibliographystyle{ieee_fullname}
\bibliography{main}
}

\appendix
\section{Outputting conditional probabilities with HXE}
\label{sec:hxe_vs_yolo}

We investigated whether outputting conditional probabilities instead of class probabilities affects the performance of the classifier represented by our proposed HXE approach (Sec.~\textcolor{red}{3.1}).
These two options correspond, respectively, to implementing hierarchical information as an architectural change or as modification of the loss only:
\begin{itemize}
	\item When the model outputs the conditional probabilities $p(C^{(l)}|C^{(l+1)})$ directly, the output dimension is equal to the total number of nodes in the hierarchy and normalisation is ensured hierarchically with a different softmax at each independent level of the hierarchy.
			Eqn.~\textcolor{red}{4} can be used directly on the output of the model.
	\item When the model outputs the class probabilities, the output dimension is equal to the number of leaf nodes in the hierarchy and normalisation can be performed with a single softmax.
			In this case, however, it is necessary to use Eqn.~\textcolor{red}{3} in order to obtain the conditional probablilities in terms of the final probabilities in Eqn.~\textcolor{red}{4}.
\end{itemize}
The second method, which we advocate here, has the advantage that hierarchichal information is implemented in the loss only, meaning that it does not require direct knowledge of the hierarchy for inference.

\begin{figure*}[!h]
	\centering
	\includegraphics[width=0.33\textwidth]{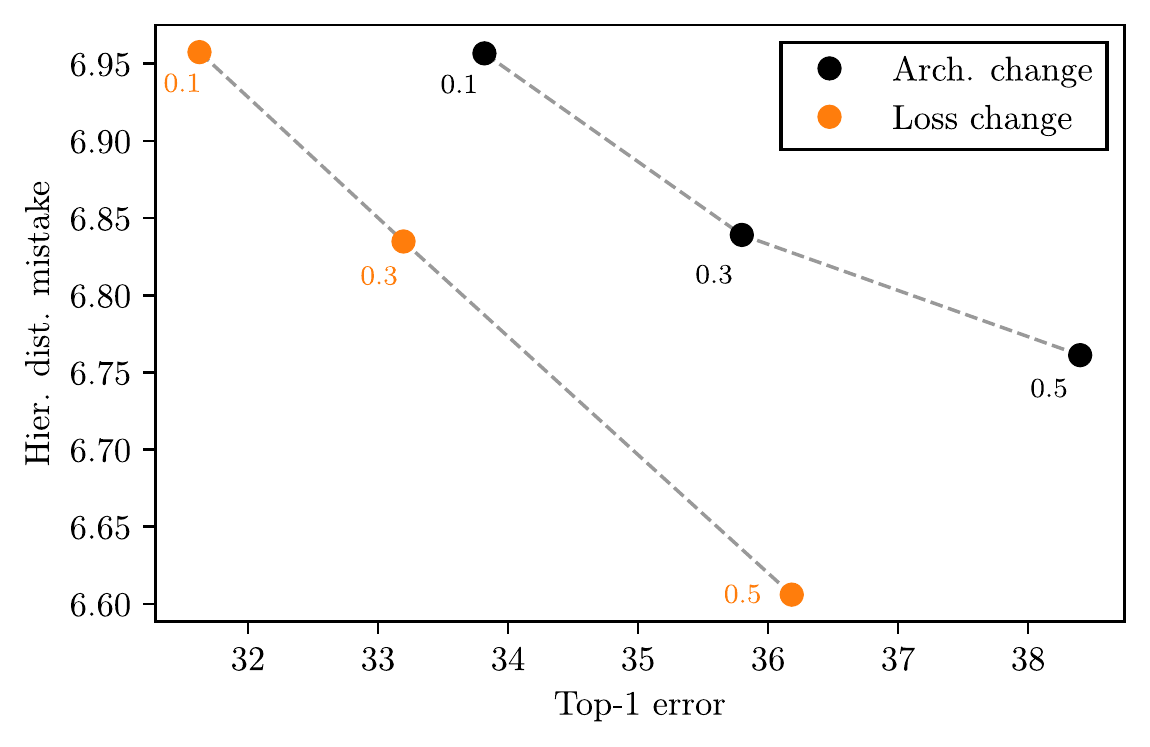}
	\includegraphics[width=0.33\textwidth]{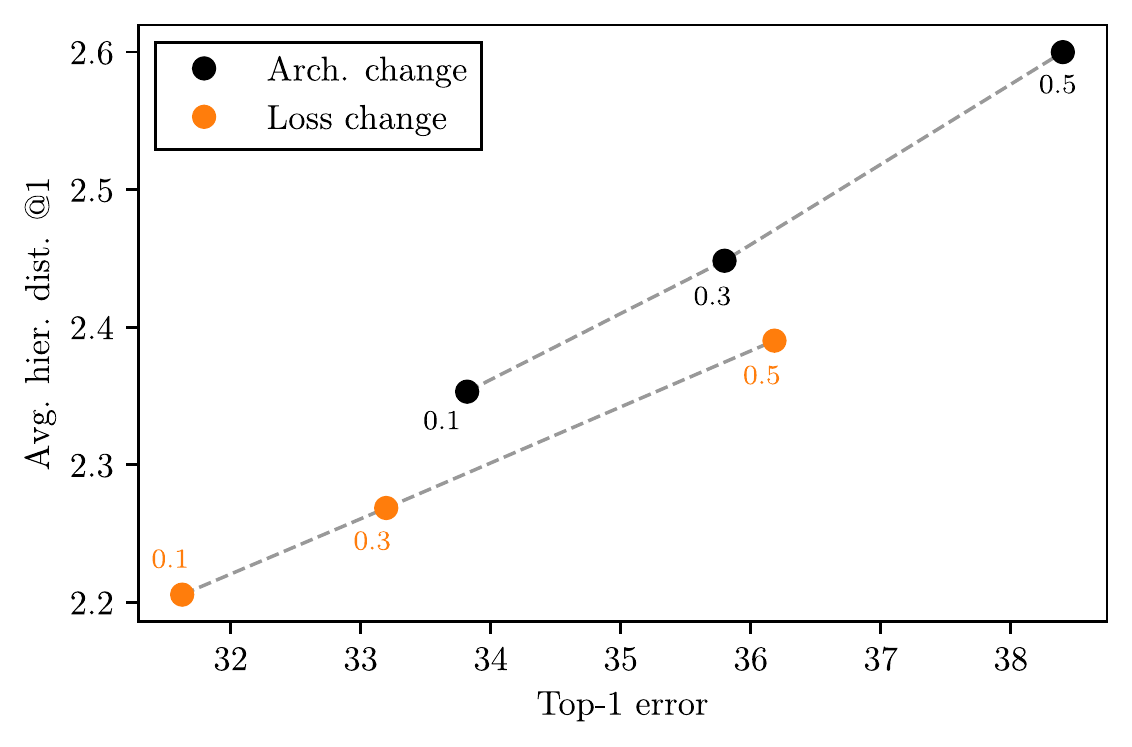}
	\includegraphics[width=0.33\textwidth]{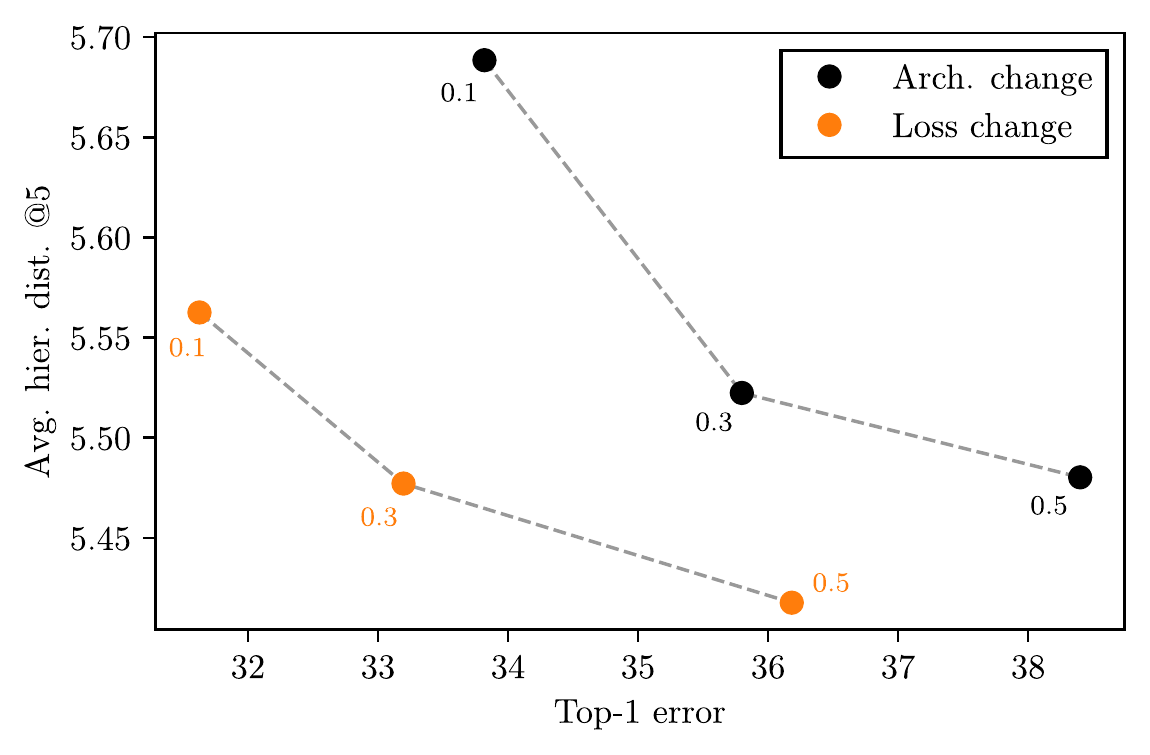} \\
	\includegraphics[width=0.33\textwidth]{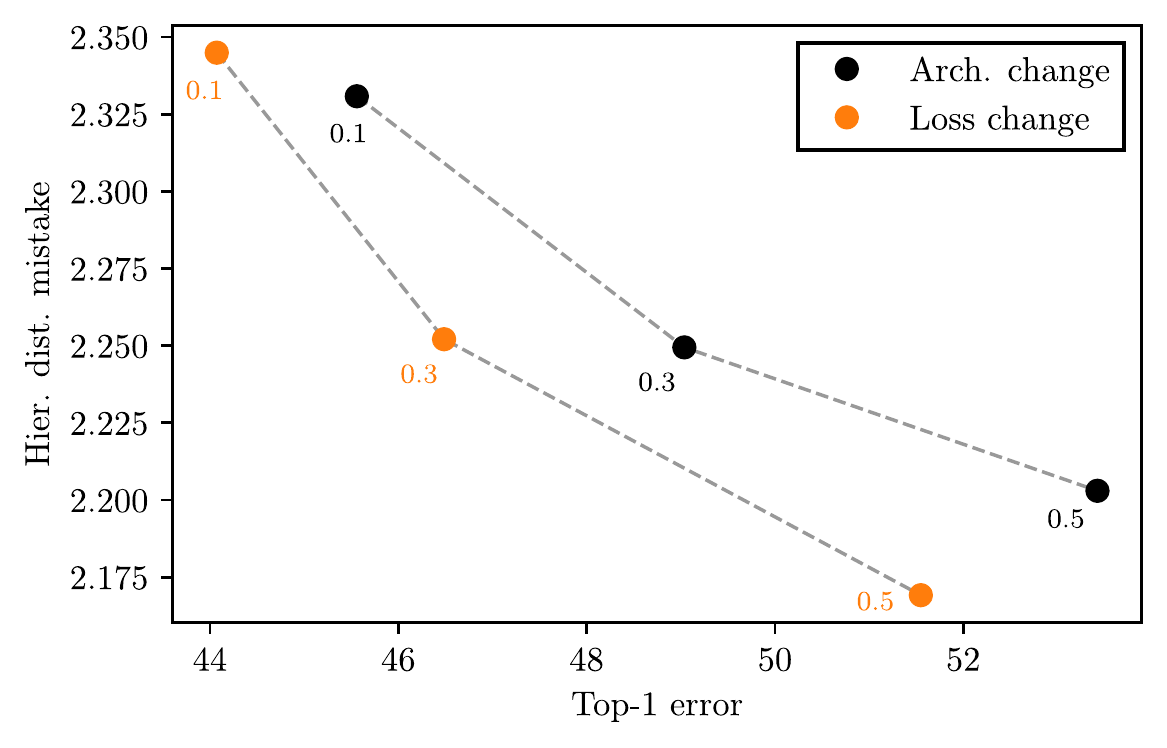}
	\includegraphics[width=0.33\textwidth]{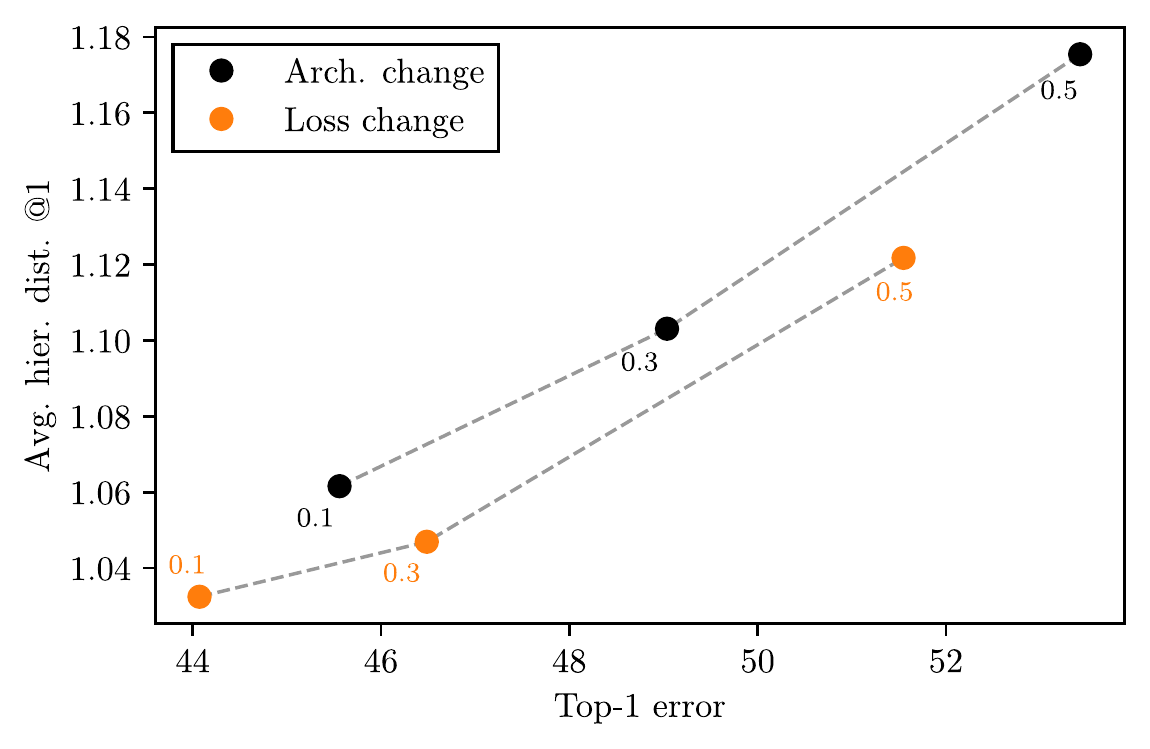}
	\includegraphics[width=0.33\textwidth]{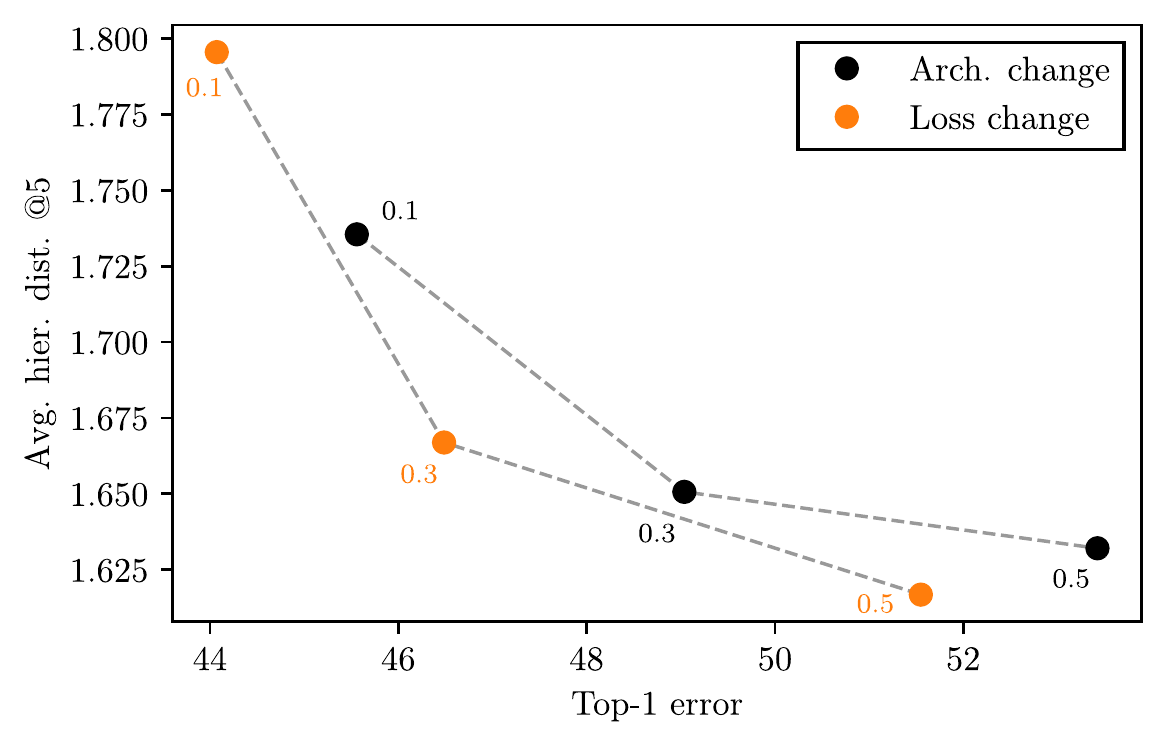}
	\captionof{figure}{\label{fig:yolo-vs-hxe}
		Outputting the conditional probabilities (architectural change) results in a degradation of performance compared to outputting the class probabilities directly (loss change) when using the hierarchical cross-entropy loss with exponential weights $\lambda(C) = \exp(-\alpha h(C))$.
		Results are shown both on \emph{tieredImageNet-H} (top) and \emph{iNaturalist19-H} (bottom).
		Points closer to the bottom-left corner of the plots are the ones achieving the best tradeoff.}
\end{figure*}

Comparing different values of $\alpha$ for otherwise identical training parameters, we also observe that outputting the class probabilities consistently results in an improvement of performance across \emph{all} of the metrics: see Fig.~\ref{fig:yolo-vs-hxe}.

\section{Note on methods based on hierarchical architectures}
\label{sec:note_hier_arch}
We opted not to evaluate against the ``generalist/expert" hierarchical models surveyed in Sec.~\textcolor{red}{2.3} for several reasons. 
For one, none of the listed methods perform experiments under hierarchical measures.
As noted in the main text, they all rely on discovered hierarchies, ruling out direct comparison to methods accepting the hierarchies considered here.
Crucially, these methods increase the capacity of their base models, which not only rules out controlled experimental comparison, but confounds the intuition behind why their designs demonstrate improved performance: the extent to which the hierarchical design per se is responsible for the gains in top-$k$ accuracy observed is actually an open question.
We also note that a recurring theme in these works is the observation that in practice, the use of generalists which make hard categorisations into disjoint coarse categories causes enough irrecoverable errors to motivate moving away from this design strategy.
Mitigating approaches typically involve the use of non-disjoint coarse categories and probabilistic one-to-many coarse classification.
Thus, while these methods are worthy of mention and further investigation, they do not yet represent attempts at the problem we examine in this paper.

\section{More implementation details}
\label{sec:more_details}

In order to perform meaningful comparisons, we adopted a simple configuration (network architecture, optimiser, data augmentation, \dots) and used it for all the methods presented in this paper.

We used a ResNet-18 architecture (with weights pretrained on ImageNet) trained with Adam~\cite{reddi2019convergence} for \num[group-separator={,}]{200000} steps and mini-batch size of \num{256}.
We used a learning rate of $1\mathrm{e}{-5}$ unless specified otherwise.
To reduce overfitting, we adopted PyTorch's basic data augmentation routines with default hyperparameters: \texttt{RandomHorizontalFlip()} and \texttt{RandomResizedCrop()}.
For both datasets, images have been resized (stretched) to $224{\times}224$.

Below, we provide further information about the methods we compared against, together with the few minor implementation choices we had to make.
As mentioned in Sec.~\textcolor{red}{1}, these methods represent, to the best of our knowledge, the only modern attempts to deliberately reduce the semantic severity of a classifier's mistakes that are generally applicable to any modern architecture.

\smallpar{\yolo.}
In motivating the hierarchical variant of the \yolo{} framework, Redmon \& Farhadi~\cite[Sec.~4]{redmon2017yolo9000}, mention the need of integrating the smaller COCO detection dataset~\cite{lin2014microsoft} with the larger ImageNet classification dataset under a unified class hierarchy.
Their approach too relies on a heuristic for converting the WordNet graph into a tree, and then effectively training a conditional classifier at every parent node in the tree by using one softmax layer per sibling group and training under the usual softmax loss over leaf posteriors.
The authors report only a marginal drop in standard classification accuracy when enforcing this tree-structured prediction, including the additional internal-node concepts.
They note that the approach brings benefits, including graceful degradation on new or unknown object categories, as the network is still capable of high confidence in a parent class when unsure as to which of its children is correct.

Since the model outputs conditional probabilities instead of class probabilities, we changed the output dimension of the terminal fully-connected layer, such that it outputs logits for every node in the hierarchy.
Proper normalisation of the conditional probabilities is then enforced at every node of the hierarchy using the softmax function. 
Finally, the loss is computed by summing the individual cross-entropies of the conditional probabilities on the path connecting the ground-truth label to the root of the tree.

\smallpar{\devise.}
\begin{figure}[t]
\lstset{language=Python}
\lstset{frame=lines}
\lstset{caption={Network head used for \devise{}.}}
\lstset{label={lst:devise}}
\lstset{basicstyle=\footnotesize}
\begin{lstlisting}
model.fc = torch.nn.Sequential(
  torch.nn.Linear(in_features=512,
                  out_features=512),
  torch.nn.ReLU(),
  torch.nn.BatchNorm1d(512),
  torch.nn.Linear(in_features=512,
                  out_features=300)
)
\end{lstlisting}
\end{figure}
Frome~\etal~\cite{frome2013devise} proposed DeViSE with the aim of both making more semantically reasonable errors and enabling zero-shot prediction.
The approach involves modifying a standard deep classification network to instead output vectors representing semantic embeddings of the class labels.
The label embeddings are learned through analysis of unannotated text~\cite{mikolov2013efficient} in a separate step, with the classification network modified by replacing the softmax layer with a learned linear mapping to that embedding space.
The loss function is a form of ranking loss which penalises the extent of greater cosine similarity to negative examples than positive ones.
Inference comprises finding the nearest class embedding vectors to the output vector, again under cosine similarity.

Since an official implementation of \devise{} is not available to the public, we re-implemented it following the details discussed in the paper~\cite{frome2013devise}.
Below the list of changes we found appropriate to make.
\begin{tight_it}
    \item For the generation of the word embeddings, instead of the rather dated method of Mikolov~\etal~\cite{mikolov2013efficient}, we used the high-performing and publicly available\footnote{\url{https://github.com/facebookresearch/fastText}} fastText library~\cite{bojanowski2017enriching} to obtain word embeddings of length 300 (the maximum made available by the library).
    \item Instead of a single fully-connected layer mapping the network output to the word embeddings, we used the network ``head'' described in Listing~\ref{lst:devise}.
		  We empirically verified that this configuration with two fully-connected layers outperforms the one with a single fully-connected layer.
		  Moreover, in this way the number of parameters of \devise{} roughly matches the one of the other experiments, which have architectures with a single fully-connected layer but a higher number of outputs (608, equivalent to the number of classes of \tih{}, as opposed to 300, the word-embedding size).
    \item Following what described in~\cite{frome2013devise}, we performed training in two steps.
		  First, we trained only the fully-connected layers for the first \num[group-separator={,}]{150000}  steps with a learning rate of $1\mathrm{e}{-4}$.
		  We then trained the entire network for \num[group-separator={,}]{50000} extra epochs, using a learning rate of $1\mathrm{e}{-6}$ for the weights of the backbone.
		  Note that~\cite{frome2013devise} did not specify neither how long the two steps of training should last nor the values of the respective learning rates.
		  To decide the above values, we performed a small hyperparameter search.
	  \item \cite{frome2013devise} says that \devise{} is trained starting from an ImageNet-pretrained architecture.
		  Since we evaluated all methods on \tih{}, we instead initialised \devise{} weights with the ones of an architecture fully trained with the cross-entropy loss on this dataset.
		  We verified that this obtains better results than starting training from ImageNet weights.
\end{tight_it}

\smallpar{\BD~\cite{barz2019hierarchy}.}
This approach involves first mapping class labels into a space in which dot products represent semantic similarity (based on normalised LCA height), then training a deep network to learn matching feature vectors (\emph{before} the fully connected layer) on its inputs.
There is a very close relationship to \devise{}~\cite{frome2013devise}, with the main difference being that here, the label embedding is derived from a supplied class hierarchy in a straightfoward manner instead of via text analysis: iterative arrangement of embedding vectors such that all dot products equal respective semantic similarities.
The authors experiment with two different loss functions: (1) a linear reward for the dot product between the output feature vector and ground-truth class embedding (\ie~a penalty on misalignment); and (2) the sum of the preceding and a weighted term of the usual cross-entropy loss on the output of an additional fully connected layer with softmax.
We only used (2), since in~\cite{barz2019hierarchy} it attains significantly better results than (1).

We used the code released by the authors \footnote{\url{https://github.com/cvjena/semantic-embeddings}} to produce the label embeddings.
To ensure consistency with the other experiments, two differences in implementation with respect to the original paper were required.
\begin{tight_it}
\item We simply used a ResNet-18 instead of the architectures Barz \& Denzler experimented with in their paper~\cite{barz2019hierarchy} (\ie ResNet-110w~\cite{he2016deep}, PyramidNet-272-200~\cite{han2017deep} and Plain-11~\cite{barz2018deep}).
\item Instead of SGD with warm restarts~\cite{loshchilov2016sgdr}, we used Adam~\cite{reddi2019convergence} with a learning rate of $1\mathrm{e}{-4}$ (the value performing best on the validation set).
\end{tight_it}

\section{Pruning the WordNet hierarchy}
\label{sec:prune_wordnet}

The ImageNet dataset~\cite{russakovsky2015imagenet} was generated by populating the WordNet~\cite{miller1998wordnet} hierarchy of nouns with images.
WordNet is structured as a graph composed of a set of IS-A parent-child relationships.
Similarly to the work of Morin \& Bengio~\cite{morin2005hierarchical} and Redmon \& Farhadi~\cite{redmon2017yolo9000}, our proposed hierarchical cross entropy loss (HXE, Sec.~\textcolor{red}{3.1}) also relies on the assumption that the hierarchy underpinning the data takes the form of a tree.
Therefore, we modified the hierarchy to obtain a tree from the WordNet graph.

First, for each class, we found the longest path from the corresponding node to the root.
This amounts to selecting the paths with the highest discriminative power with respect to the image classes.
When multiple such paths existed, we selected the one with the minimum number of new nodes and added it to the new hierarchy.
Second, we removed the few non-leaf nodes with a single child, as they do not possess any discriminative power.

Finally, we observed that the pruned hierarchy's root is not \textsc{physical entity}, as one would expect, but rather the more general \textsc{entity}.
This is problematic, since \textsc{entity} contains both physical objects \emph{and} abstract concepts, while \emph{tieredImageNet} classes only represent physical objects.
Upon inspection, we found that this was caused by the classes \textsc{bubble}, \textsc{traffic sign}, and \textsc{traffic lights} being connected to \textsc{sphere} and \textsc{sign}, which are considered abstract concepts in the WordNet hierarchy.
Instead, we connected them to \textsc{sphere}, \textsc{artifact} and \textsc{signboard}, respectively, thus connecting them to \textsc{physical entity}.

Even though our second proposed method (\emph{soft labels}), as well as the cross-entropy baseline, \devise{}~\cite{frome2013devise} and Barz \& Denzler~\cite{barz2019hierarchy}, do not make any assumption regarding the structure of the hierarchy, we ran them using this obtained pruned hierarchy for consistency of the experimental setup.

\clearpage
\onecolumn
\section{Supplementary figures}

\begin{figure*}[!h]
	\centering
	\includegraphics[width=0.5\textwidth]{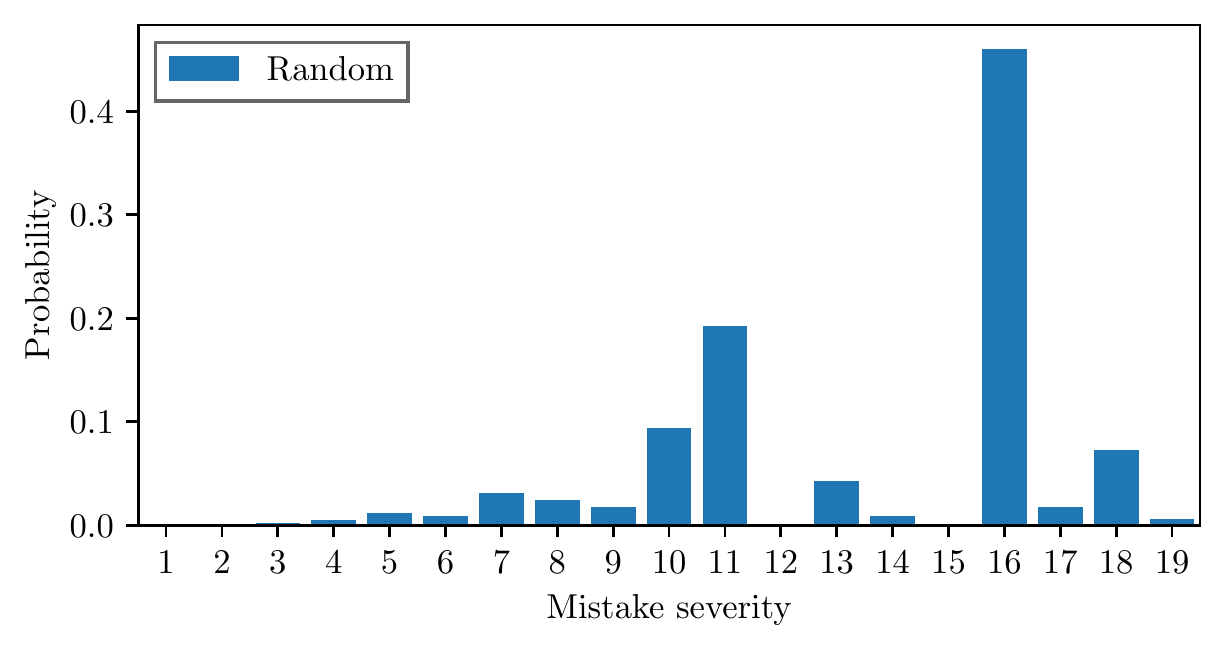}
	\caption{\label{fig:random hist}
		Distribution of mistake severity when drawing random pairs of images from  ImageNet/ILSVRC-12.
		Note that though this distribution shares some qualitative similarities with the ones shown in Fig.~\textcolor{red}{1}, it is nonetheless substantially different.
		This suggests that the shapes of the mistake-severity distributions for the various DNN architectures studied cannot be explained by properties of the dataset alone.}
\end{figure*}

\end{document}